%% file: main.tex
\documentclass[runningheads]{llncs}
\usepackage{cite}
\usepackage{amsmath,amssymb,amsfonts}
\usepackage{algorithmic}
\usepackage{graphicx}
\usepackage{textcomp}
\usepackage{xcolor}
\usepackage{xspace}
\newcommand{\tool}{\textit{STAN}\xspace}
\newcommand{\toola}{\textit{STAN-a}\xspace}
\newcommand{\toolb}{\textit{STAN-b}\xspace}
\usepackage{color}
\usepackage{multirow}
\usepackage{amsfonts}
\usepackage{mathtools}
\usepackage{algorithm}
\usepackage{algorithmic}
\usepackage{tcolorbox}
\usepackage{subcaption}
\usepackage{graphicx}
\usepackage{booktabs}
\usepackage{varwidth}
\usepackage{tabularx}
\usepackage{tcolorbox}
\usepackage{tabularx}
\usepackage[switch]{lineno}  %
\usepackage{makecell}
\newcommand{\linebreakand}{%
  \end{@IEEEauthorhalign}
  \hfill\mbox{}\par
  \mbox{}\hfill\begin{@IEEEauthorhalign}
}
\usepackage{graphicx}

\begin{document}
\title{STAN: Synthetic Network Traffic Generation\\with Generative Neural Models
}
\titlerunning{STAN}
%

\author{Shengzhe Xu\inst{1} \and Manish Marwah\inst{2} \and Martin Arlitt\inst{2} \and Naren Ramakrishnan\inst{1}}

%
%
\institute{Department of Computer Science, Virginia Tech, Arlington, VA, USA \email{shengzx@vt.edu, naren@cs.vt.edu}\and Micro Focus, Santa Clara, CA, USA \email{\{manish.marwah,martin.arlitt\}@microfocus.com}}
%
\maketitle              
%

%
%
%
\input{abstract}

\input{intro}

\input{relatedwork.tex}

\input{problemdefinition.tex}

\input{method.tex}

\input{model.tex}

\input{evaluation.tex}

\input{conclusion.tex}

%
%

\input{main.bbl}

\end{document}

%% file: abstract.tex
\begin{abstract}
Deep learning models have achieved great success in recent years but progress in some domains like cybersecurity is stymied
due to a paucity of realistic datasets.
Organizations are reluctant to share such data, even internally, due to privacy reasons. An alternative is to use synthetically generated data but existing methods are limited
in their ability to capture complex dependency structures, between attributes
and across time. This paper presents \tool (Synthetic network Traffic generation with Autoregressive Neural models), a tool to generate realistic synthetic network traffic datasets for subsequent downstream
applications. Our novel neural architecture captures both temporal dependencies and dependence between attributes at any given time. It integrates convolutional neural layers with mixture density neural layers and softmax layers, and models both continuous and discrete variables. We evaluate the performance of \tool in terms of quality of data generated, by training it on both a simulated dataset and a real network traffic data set. 
Finally, to answer the question---can real network traffic data be substituted with synthetic data to train models of comparable accuracy?---we train two anomaly detection models based on self-supervision.
The results show only a small decline in accuracy of models trained solely on synthetic data. 
While current results are encouraging in terms of quality of data generated and absence of any obvious data leakage from  training data, in the future we plan to further validate this fact by conducting privacy attacks on the generated data. Other future work includes validating capture of long term dependencies and making model training more efficient. 
\end{abstract}

%% file: intro.tex
\section{Introduction}

Cybersecurity has become a key concern for both
private and public organizations, given the prevalence of cyber-threats and attacks.
In fact, malicious cyber-activity cost the U.S. economy
between \$57 billion and \$109 billion in 2016 \cite{whitehousereport}, and worldwide
yearly spending on cybersecurity reached \$1.5 trillion in 2018 \cite{riskiq2019}.

To gain insights into and to counter cybersecurity threats, organizations
need to sift through large amounts of network, host, and
application data typically produced in an organization.  Manual inspection of
such data by security analysts to discover attacks is impractical due
to its sheer volume, e.g., even a medium-sized organization can
produce terabytes of network traffic
data in a few hours. Automating the process through use of machine
learning tools is the only viable alternative.  Recently, deep learning
models have been successfully used for cybersecurity applications~\cite{berman2019survey,kwon2017survey}, and
given the large quantities of available data, deep learning methods appear
to be a good fit.

However, although large amounts of data is available to security teams for cybersecurity
machine learning applications, it is sensitive in nature and access to it
can result in privacy violations, e.g.,
network traffic logs can reveal web browsing behavior of users. Thus,
it is difficult for data scientists to obtain realistic data to train their models,
even internally within an
organization. To address data privacy issues,
three main approaches are usually explored~\cite{Aggarwal2008,al2019privacy}: 1) non-cryptographic anonymization methods; 2) cryptographic anonymization methods; and, 3) perturbation methods, such as differential privacy. However, 1) leaks private information
in most cases; 2) is currently impractical (e.g., using homomorphic encryption) for large data sets; and, 3) degrades data quality
making it less suitable for machine learning tasks. 

\begin{figure}[t]
\centering
\includegraphics[width=\linewidth]{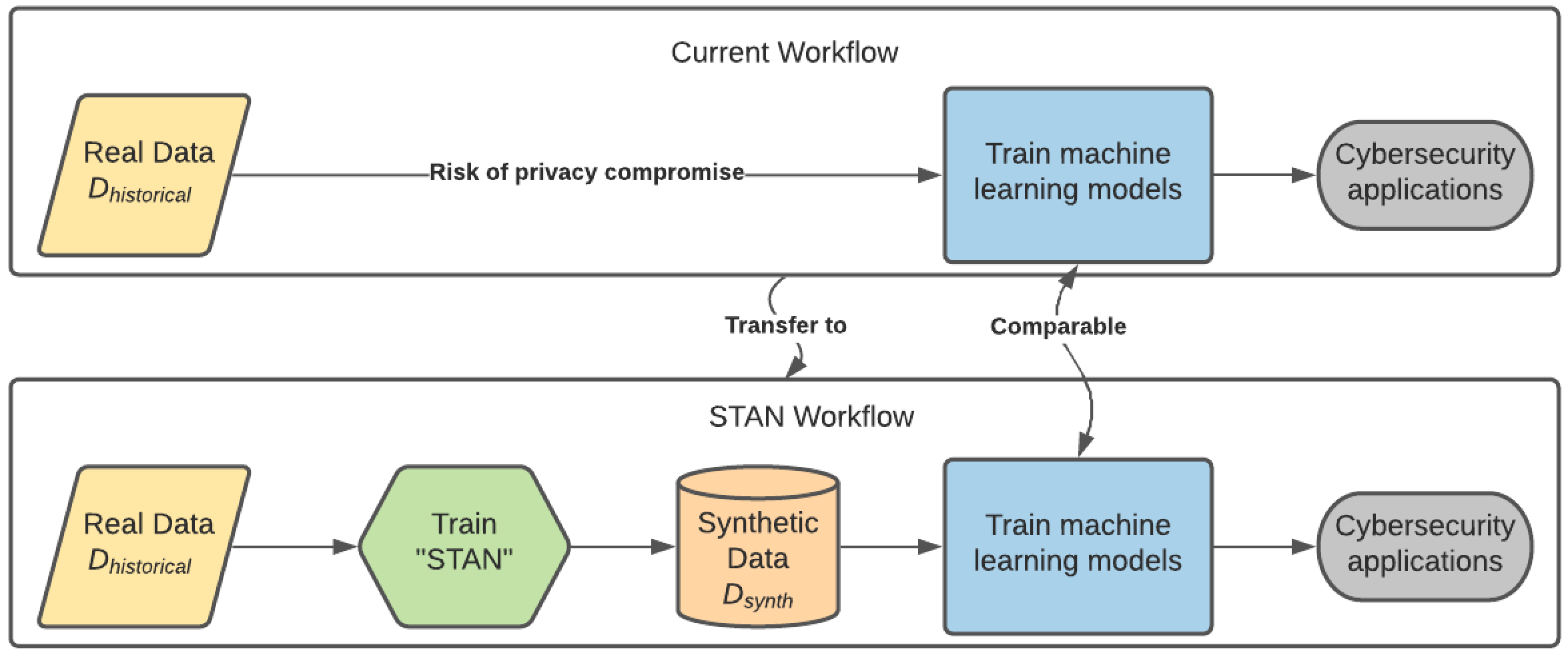}
\caption{\tool The top figure shows a simplified workflow where real data $(D_{historic})$ is used to train a machine learning model for cybersecurity applications; however, use of real data may result in a privacy compromise. The bottom figure shows the proposed workflow, where machine learning models are trained using realistic synthetic data $(D_{synth})$ generated by STAN.  
}
\label{fig:overview}
\end{figure}

In this paper, we take an orthogonal approach. We generate synthetic data that is realistic enough to replace real data in machine learning tasks. Specifically, 
we consider multivariate time-series data and, unlike prior work, capture both temporal dependencies and dependencies between attributes.
Figure~\ref{fig:overview} illustrates our approach, called \tool.
Given real historical data, we first train a CNN-based autoregressive generative neural network that learns the underlying joint data distribution.
The trained STAN model can then generate any amount of synthetic data without revealing private information.
This realistic synthetic data can replace real data in the training of machine learning models applied to cybersecurity applications, with performance\footnote{Here performance refers to a model evaluation metric such as precision, recall, F1-score, mean squared error, etc.} comparable to models trained on real data.

To evaluate the performance of \tool, we use both simulated data and a real publicly-available network traffic data set.
We compare our method with four selected baselines using several metrics to evaluate the quality of the generated data.
We also evaluate the suitability of the generated data as compared to real data in training machine learning models. 
Self-supervision is commonly used for anomaly detection \cite{vyas2018out, chen2018real}.
We consider two such tasks
-- a classification task and a regression task for detecting
cybersecurity anomalies -- that are trained on both real and synthetic data. 

We show comparable model performance after entirely substituting the real training data with our synthetic data: the F-1 score of the classification task only drops by 2\% (99\% to 97\%), 
while the mean square error only increases by about 13\% for the regression task.

In this paper, we make the following key contributions:

\begin{itemize}
  \item \tool is a new approach to learn joint distributions of
  multivariate time-series data---data typically used in cybersecurity applications---and then to generate synthetic data from the learned distribution. Unlike prior work, \tool learns both temporal and attribute dependencies. It integrates convolutional neural layers (CNN) with mixture density neural layers (MDN) and softmax layers to model both continuous and discrete variables. Our code is publicly available.\footnote{https://github.com/ShengzheXu/stan.git}
\item  We evaluated \tool on both simulated data and a real publicly available network traffic data set, and compared with four baselines. 
\item We built models for two cybersecurity machine learning tasks using only \tool generated data to train and which demonstrate model performance comparable
to using real data.   
\end{itemize}

%% file: relatedwork.tex
\section{Related Work}

\textbf{Synthetic data generation.} Generating synthetic data to make up for the lack of real data is a common solution. Compared to modeling image data \cite{oord2016pixel}, learning distributions on multi-variate time-series data poses a different set of challenges. Multi-variate data is more diverse in the real world, and such data usually has more complex dependencies (temporal and spatial) as well as heterogeneous attribute types (continuous and discrete).

Synthetic data generation models often treat each column as a random variable to model joint multivariate probability distributions. The modeled distribution is then used for sampling. Traditional modeling algorithms~\cite{avino2018generating,graham2012differentially,sun2019learning} have the restraint of distribution data types and due to computational issues, the dependability of synthetic data generated by these models is extremely limited. Recently, GAN-based (Generative Adversarial Network-based) approaches augment performance and flexibility to generate data~\cite{park2018data,xu2019modeling,lin2020using}.

However, they are still either restricted to a static dependency without considering the temporal dependence usually prevalent in real world data~\cite{ring2019flow, xu2019modeling}, or only partially build temporal dependence inside GAN blocks~\cite{jan2020throwing, lin2020using}. By `partially' here, we mean that the temporal dependencies generated by GANs is based on RNN/LSTM decoders, which limits the temporal dependence in a pass or a block that GAN generates at a time. In real \textit{netflow} data, traffic is best modeled as infinite flows which should not be interrupted by blocks of the synthesizer model. Approaches such as~\cite{jan2020throwing} (which generates embedding-level data, such as the embedding of URL addresses) and \cite{lin2020using} are unable to reconstruct real \textit{netflow} data characteristics such as IP address or port numbers, and other complicated attribute types.
We are not aware of any prior work that models both entire temporal and between-attribute dependencies for infinite flows of data, as shown in Table \ref{table:arts}.

Furthermore, considering the \textit{netflow} domain speciality, such data should not be simply treated as traditional tabular data. For example, \cite{ring2019flow} passes the \textit{netflow} data into a pre-trained IP2Vec encoder \cite{ring2017ip2vec}  to obtain a continuous representation of IP address, port number, and protocol for the GAN usage. \tool doesn't need a separate training component, and instead successfully learns IP address, port number, protocol, TCP flags characteristics naturally by simply defining the formats of those special attributes.

\begin{table*}[ht]
\centering
\begin{tabular}{l | c| c| c}
\toprule
Methods & \makecell[l]{Captures \\attribute \\dependence} & \makecell[l]{Captures \\temporal \\dependence} & 
\makecell[l]{Generates\\ realistic IP addresses/ \\ port numbers } \\
\midrule
WP-GAN \cite{ring2019flow} & \checkmark & & \checkmark\\
CTGAN \cite{xu2019modeling} & \checkmark & & \\
ODDS \cite{jan2020throwing} & \checkmark& partially &  \\
DoppelGANger \cite{lin2020using} &\checkmark& partially&\\
\midrule
\tool      & \checkmark    & \checkmark & \checkmark  \\
\bottomrule
\end{tabular}
\caption{Comparison of STAN with recent similar work}
\label{table:arts}
\end{table*}

\textbf{Autoregressive generative models.} \cite{van2016conditional,oord2016pixel} have been successfully applied to signal data, image data, and natural language data. They attempt to iteratively generate data elements: previously generated elements are used as an input condition for generating the subsequent data.
Compared to GAN models, autoregressive models emphasize two factors during the distribution estimating: 1) the importance of the time-sequential factor; 2) an explicit and tractable density. In this paper, we apply the autoregressive idea to learn and generate time-series multi-variable data.

\textbf{Mixture density networks.} Unlike modeling discrete attributes, some continuous numeric attributes are relatively sparse and span a large value range. Mixture Density Networks~\cite{bishop1994mixture} is a neural network architecture to learn a Gaussian mixture model (GMM) that can predict continuous attribute distributions. 
This architecture provides the possibility to integrate GMM into a complex neural network architecture.

\textbf{Machine learning for cybersecurity.} In the past decades, machine learning has been brought to bear upon multiple tasks in cybersecurity, such as automatically detecting malicious activity and stopping attacks~\cite{catania2012automatic,buczak2015survey}. Such machine learning approaches usually require a large amount of training data with specific features. However, training model using real user data leads to privacy exposure and ethics problems. Previous work on anonymizing real data \cite{razak2020data} has failed to provide satisfactory privacy protection, or degrades data quality too much for machine learning model training.

This paper takes an different approach that, by learning and generating realistic synthetic data, ensures that real data can be substituted when training machine learning models.

Prior work on generating synthetic network traffic data includes Cao et al.
\cite{cao2004stochastic} who use a network simulator to generate traffic data while we do not generate data through workload or network simulation;  both Riedi et al. and Paxson \cite{riedi1999multifractal, paxson1997fast} use advanced time-series methods, however, these are for univariate data, and different attributes are assumed to be independent. We model multivariate data, that is, all attributes at each time step, jointly. 
Mah \cite{mah1997empirical} models distributions of HTTP packet variables similar to baseline 1 (B1) in our paper. Again, unlike our method, it does not jointly model the network data attributes.

%% file: problemdefinition.tex
\section{Problem Definition}

\begin{figure*}[t]
\centering
\includegraphics[width=\textwidth]{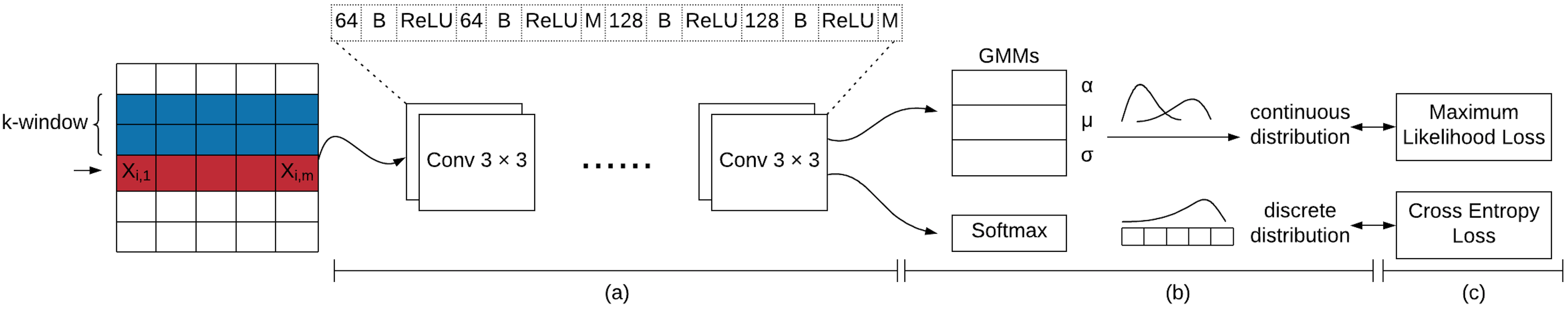}
\caption{\tool components: (a) window CNN, which crops the context based on a sliding window and extracts features from context; The CNN architecture includes 14 layers where numeric values notes are the number of 3*3 convolutional filters; B notes batch normalization layers; ReLU notes activation layers; and M notes max pooling layers.  (b) mixture density neural layers and softmax layers learn to predict the distributions of various types of attributes; (c) the loss functions for different kinds of layers.}
\label{fig:nn_struct}
\end{figure*}

We assume the data to be generated is a multivariate
time-series. Specifically, data set $\textbf{x}$ contains \textbf{$n$}
rows and \textbf{$m$} columns. 
Each row $\textbf{x}_{(i,:)}$ is an observation at time point $i$ and 
each column $\textbf{x}_{(:,j)}$ is a random
variable $j$, where $i \in [1..n]$ and $j \in [1..m]$.
Unlike typical tabular data, e.g., found in relational database
tables, and unstructured data, e.g., images, multivariate time-series
data poses two main challenges: 1) the rows are
generated by an underlying temporal process and are thus not
independent, unlike traditional tabular data; 2) the columns or attributes are not necessarily
homogeneous, and comprise multiple data types such as  numerical,
categorical or continuous, unlike say images.

The data $\textbf{x}$ follows an unknown, high-dimensional joint distribution
$\mathbb{P}$(\textbf{x}), which is infeasible to estimate directly.
The goal is to estimate 
$\mathbb{P}$(\textbf{x}) by a generative model $\mathbb{S}$ which retains the dependency structure across rows and columns. Values in a column typically depend on other columns, and temporal dependence of a row can extend to many prior rows. 
Once model $\mathbb{S}$ is trained, it can be used to generate an arbitrary amount of data, $\textbf{D}_{synth}$.

Another key challenge is evaluating the quality of the generated data,
$\textbf{D}_{synth}$.
Assuming a data set, $\textbf{D}_{historical}$, is used to
train $\mathbb{S}$, and an unseen test data set,  $\textbf{D}_{test}$, is
used to evaluate the performance of $\mathbb{S}$,
we use two criteria to compare $\textbf{D}_{synth}$ with
$\textbf{D}_{test}$:
\begin{enumerate}

    \item Similarity between a metric $M$ evaluated on the two data sets, that is, is $M(\textbf{D}_{test}) \approx M(\textbf{D}_{synth})?$
    
    \item Similarity between performance $P$ on training the same machine learning task $T$, in which the real data, $\textbf{D}_{test}$, is replaced by the synthetic data, $\textbf{D}_{synth}$, that is, is
    $P[T(\textbf{D}_{test})] \approx P[T(\textbf{D}_{synth})]?$
\end{enumerate}

%% file: method.tex
\section{Proposed Method}
\label{sec:method}

We model the joint data distribution, $\mathbb{P}$(\textbf{x}), using an autoregressive neural network.

The model architecture, shown in Figure \ref{fig:nn_struct}, combines CNN layers with a density mixture network \cite{bishop1994mixture}. The CNN captures temporal and spatial (between attributes) dependencies, while the density mixture neural layer uses the learned representation to model the joint distribution. More architecture details will be discussed in Section \ref{section_nnarch}.
During the training phase, for each row, \tool takes a data window prior to it as input. Given this context, the neural network learns the conditional distribution for each attribute. 
Both continuous and discrete attributes can be modeled. While a density mixture neural layer is used for continuous attributes, a softmax layer is used for discrete attributes. 


In the synthesis phase, \tool sequentially generates each attribute in each row. Every generated attribute in a row, having been sampled from a conditional distribution over the prior context, serves as the next attribute's context.

\subsection{Joint distribution factorization}

$\mathbb{P}$(\textbf{x}) denotes the joint probability of data $\textbf{x}$ composed of \textbf{$n$} rows and \textbf{$m$}  attributes. 
We can expand the data as a one-dimensional sequence $\textbf{x}_{1},...,\textbf{x}_{n}$, where each vector $\textbf{x}_i$ represents one row including the $m$ attributes $x_{i,1},...,x_{i,m}$. To estimate the joint distribution $\mathbb{P}$(\textbf{x}) we express it as the product of conditional distributions over the rows.
We start from the joint distribution factorization with no assumptions:

\begin{equation}
\label{jd:pixelrnn}
\mathbb{P}(\textbf{x}) = \prod_{i=1}^{n} \mathbb{P}(\mathbf{x}_i | \mathbf{x}_{1},...,\mathbf{x}_{i-1})
\end{equation}

Unlike unstructured data such as images, multivariate time-series data usually corresponds to underlying continuous processes in the real world and do not have exact starting and ending points.
It is impractical to make a 
row probability $\mathbb{P}(\textbf{x}_i)$ depend on all prior rows as in Equation \ref{jd:pixelrnn}. Thus, a $k$-sized sliding window is utilized to restrict the context to only the $k$ most recent rows. In other words, a row conditioned on the past $k$ rows is independent of all remaining prior rows, that is, 
for $i>k$, we assume independence between $\textbf{x}_i$ and $\textbf{x}_{<i-k}$. 
We can thus rewrite the joint distribution $\mathbb{P}$(\textbf{x}) as the product of the conditional distributions over the prior $k$ rows:

\begin{equation}
\label{jd:arcnn}
\mathbb{P}(\textbf{x}) = \prod_{i=1}^{k} \mathbb{P} (\mathbf{x}_i | \mathbf{x}_{1},...,\mathbf{x}_{i-1}) \prod_{i=k+1}^{n} \mathbb{P}(\mathbf{x}_i | \mathbf{x}_{i-k},...,\mathbf{x}_{i-1})
\end{equation}

Note that a suitable value of $k$ needs to be picked based on empirical evidence or domain knowledge. 
While all the probabilities in the second term on the RHS of Equation \ref{jd:arcnn} are conditioned on $k$ variables, the same is not true for the probabilities in the first term. To make these consistent, we add zero padding and
then symbolically define that $\textbf{x}_i$ where $i \leq 0$ represents a padding row, as Equation \ref{jd:arcnn-simple} shows.

\begin{equation}
\label{jd:arcnn-simple}
\begin{split}
\mathbb{P}(\textbf{x}) &= \prod_{i=1}^{k} \mathbb{P}(\mathbf{x}_i | \mathbf{x}_{i-k},..., \mathbf{x}_{1},...,\mathbf{x}_{i-1}) \prod_{i=k+1}^{n} \mathbb{P}(\mathbf{x}_i | \mathbf{x}_{i-k},...,\mathbf{x}_{i-1})\\
&= \prod_{i=1}^{n} \mathbb{P}(\mathbf{x}_i | \mathbf{x}_{i-k},...,\mathbf{x}_{i-1})
\end{split}
\end{equation}

The joint distribution of 
a row can be factorized in two ways: 1) Equation \ref{jd:arcnn-row-a} assumes conditional independence of attributes in a row, given all attributes in the previous $k$ rows; 2) Equation \ref{jd:arcnn-row-b} 
makes no conditional independence assumptions of attributes 
in the same row.

\begin{equation}
\label{jd:arcnn-row-a}
\mathbb{P}(\textbf{x}) = \prod_{i=1}^{n}\prod_{j=1}^{m} \mathbb{P}(\mathbf{x}_{i,j} | \mathbf{x}_{i-k},...,\mathbf{x}_{i-1})
\end{equation}

\begin{equation}
\label{jd:arcnn-row-b}
\mathbb{P}(\textbf{x}) = \prod_{i=1}^{n}\prod_{j=1}^{m} \mathbb{P}(\mathbf{x}_{i,j} | \mathbf{x}_{i-k},...,\mathbf{x}_{i-1}; x_{i,1},...,x_{i,j-1})
\end{equation}

While (\ref{jd:arcnn-row-a}) provides a good approximation, we found (\ref{jd:arcnn-row-b}) performs slightly better.

During generation, initialization is a key concern for our generative model due to temporal 
dependence. In order to generate data without supplying any real data or specific seed, we begin the above autoregressive chain process with the marginal distribution $\mathbb{P}(\mathbf{x}_{1})$. In practice, as described later in section (\ref{section_nnarch}), the marginal distribution is approximated by all zero conditional: $\mathbb{P}(\mathbf{x}_{1}|0)$.  Then beginning from the second step, Equation \ref{jd:arcnn-row-a}, \ref{jd:arcnn-row-b} holds.

%% file: model.tex
\subsection{Neural network architecture}
\label{section_nnarch}

As shown in Figure \ref{fig:nn_struct}, 
the input window goes through the \textit{convolutional layers} followed by \textit{mixture density neural layers} or \textit{softmax layers} sequentially to learn the joint distribution.
We define two \textit{loss} functions for the two distribution modeling layers separately. 
Algorithms \ref{alg:1} and \ref{alg:2} provide details on model training and data synthesis.
Note that the training phase allows for parallelization while the synthesis phase is sequential.

\begin{algorithm}[h]
    \caption{Model training process for each attribute $j$}
    \label{alg:1}
    \hspace*{\algorithmicindent} \textbf{Input} $D_{Historical}$, window size $k$, attribute type $T_j$. \\
    \hspace*{\algorithmicindent} \textbf{Output} STAN model $\mathbb{S}_{stan}$;
    \begin{algorithmic}[1]
    \STATE Construct window data\\
        \ \ \ \ $X^{window}_i$ = concatenate $X_{i-k}$,...,$X_{i}$;\\
        \ \ \ \ $y^{window}_i$ = $X_i$;
    
    \FOR{epoch in 1 ... EPOCH}
        \STATE $X^{window}_i$ *= $Mask$
        \IF{$T_j$ is continuous}
            \STATE $\mathbb{P}_{gmm\_pred}$ = $mdn(wCNN(X_{window}))$;\\
            \STATE $loss$ = $nll(\mathbb{P}_{gmm\_pred}, y_{window})$;\\
        \ELSE 
            \STATE $\mathbb{P}_{softmax\_pred}$ = $softmax(wCNN(X_{window}))$;\\
            \STATE $loss$ = $cross\_entropy(\mathbb{P}_{softmax\_pred}$, $y_{window})$;\\
        \ENDIF
        \STATE Using $Adam$ optimizer to update $\mathbb{S}_{stan}$ with $loss$;\\ 
    \ENDFOR
    \end{algorithmic}
\end{algorithm}

\begin{algorithm}[h]
    \caption{Data synthesis process}
    \label{alg:2}
    \hspace*{\algorithmicindent} \textbf{Input} Trained STAN model $\mathbb{S}_{stan}$. \\
    \hspace*{\algorithmicindent} \textbf{Output} $D_{synth}$;
    \begin{algorithmic}[1]
        \STATE Init context $X^{window}$ = marginal sampling()
        \WHILE{condition(target row number or time stamp)}
            \STATE $X_i^{window}$ *= $Mask$
            \STATE $P_{pred} = \mathbb{S}_{stan}(X_i^{window})$;\\
            \STATE $y_{sample}$ = sample from distribution $\mathbb{P}_{pred}$;\\
            \STATE $X_{i+1}^{window} = X_i^{window}[1:, :] + y_{sample}$
        \ENDWHILE
    \end{algorithmic}
\end{algorithm}

\textbf{Window convolutional layers (wCNN).} The CNN layers, which we
call window CNN since they operate on a sliding window of data, 
perform a two-dimensional convolution.
For one row $\textbf{x}_i$ the layers capture a rectangular context above the row as shown in Figure \ref{fig:nn_struct}
\tool uses multiple convolutional layers that preserve the spatial and temporal resolution in a sliding time window box. Each number in Figure \ref{fig:nn_struct} represents the number of $3*3$ filters in that layer. Batchnorm, ReLU and max pooling layers are also used, marked as $BN$, $ReLU$, and $M$, respectively.

\textbf{Convolution mask.}
Based on which factorization is selected, we have mask A for Equation \ref{jd:arcnn-row-a} and mask B for Equation \ref{jd:arcnn-row-b}.

\begin{figure}[ht]
\label{fig:byt_distribution}
\centering
\begin{subfigure}{.45\linewidth}
    \centering
    \includegraphics[width=\linewidth]{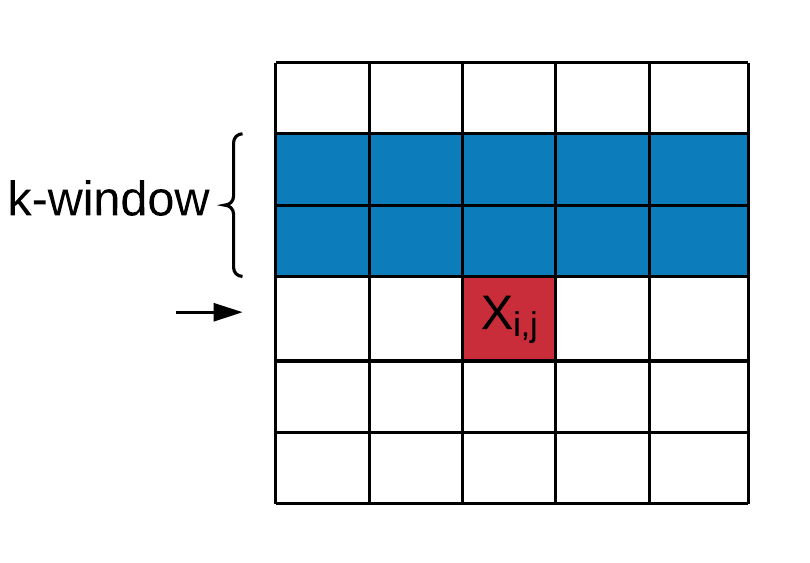}
    \subcaption{Mask A for conditional independence assumption between attributes in same row}
\end{subfigure}
\quad
\begin{subfigure}{.45\linewidth}
    \centering
    \includegraphics[width=\linewidth]{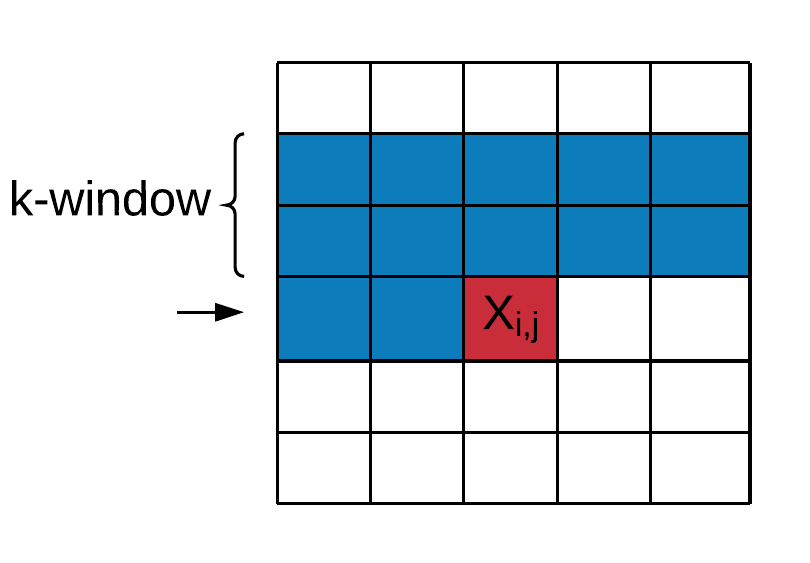}
    \subcaption{Mask B for no conditional independence assumption in the same row}
\end{subfigure}
\caption[short]{Masks for context window convolution}
\label{fig:masks}
\end{figure}

\textbf{Mixture density neural layer (mdn).}  
learns a conditional \textit{Gaussian mixture distribution}. It consists of three parallel fully connected layers, modeling $\alpha_{i}, \sigma_{i}, \mu_{i}$ separately, where the parameter $\alpha_{i}$ represents for the component weights of an \textit{Gaussian mixture model}, and the $\mu_{i}$ and $\sigma_{i}^2$ are the mean and variance parameters of the Gaussian distribution components.
The $\alpha_{i}$ parameters output go to a softmax, so that the weights of all the Gaussian mixture components sum to one.

\textbf{Loss functions.}
We define loss functions for \textit{mixture density neural layer} and \textit{softmax layer} separately.
Note that the two losses have different scales, 
and while 
multitask learning has its advantages, we match each \textit{mixture density neural component} or \textit{softmax component} with an individual \textit{wCNN component}. 

A Negative Log-Likelihood Loss (NLL) is used for the mixture density layers, which predict a group of mixture density parameters that can compose a Gaussian mixture model as Equation \ref{loss:nll}: $\alpha_{i}, \sigma_{i}, \mu_{i}$.
We use maximum likelihood loss to estimate a true distribution: the label of the input, which is the new row that to be generated, is supposed to have the highest probability in the estimated distribution. Cross entropy loss is used for the softmax layer.

\begin{equation}
NLL(x|\mu,\sigma^2) = -\log{\sum{\alpha_{i} * \mathcal{N}(x|\mu_{i}, \sigma_{i}^2)}}\label{loss:nll}
\end{equation}

\subsection{IP address and port number Modeling}
\label{sec:model_ipport}

IP addresses and port numbers are key to network traffic data. 
However, naively modeling them as continuous or discrete variables gives poor results.  

\tool specifically learns IP address and port number characteristics. 

\begin{figure}[ht]
\centering
\begin{subfigure}{\linewidth}
    \centering
    \includegraphics[width=\linewidth]{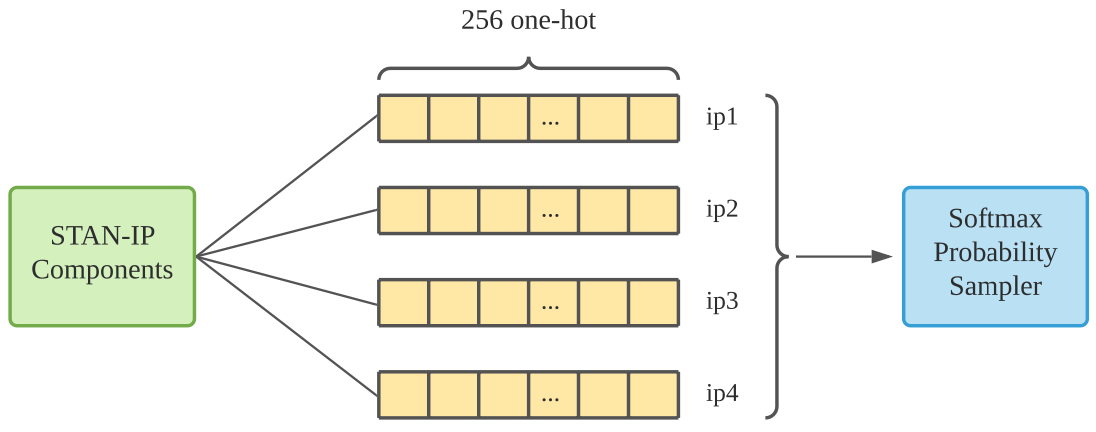}
    \subcaption{\tool IP address modeling}
\end{subfigure}
\quad
\begin{subfigure}{\linewidth}
    \centering
    \includegraphics[width=\linewidth]{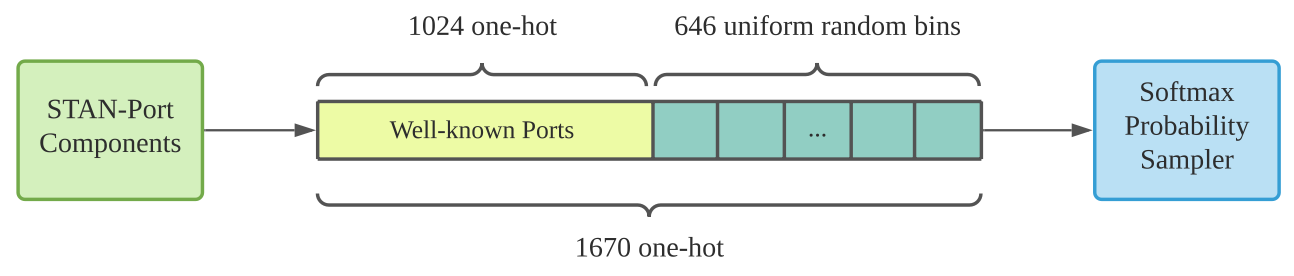}
    \subcaption{\tool port number modeling}
\end{subfigure}%
\caption[short]{With the benefit of flexible \tool continuous and discrete generator architecture, special domain attributes (such as IP address, port, protocol, and TCP flags), can be learned by purely modifying the configure parameters.}
\label{fig:ipport_arch}
\end{figure}

As shown in Figure~\ref{fig:ipport_arch} (a), 
\tool treats an IP address as four 256-categorical discrete attributes. With the help of the \tool Mask definition, even though one IP address attribute is split across four intermediate attributes, it is considered together as one variable for attribute dependence. 

Since port numbers can vary between 0 and 64K, that is too many values to model as a discrete variable. On the other hand,
treating it as a continuous variable would result in inaccuracies especially for well-known ports (those less than 1024), where being off even by one can mean something completely different. 
Therefore, we take a hybrid approach.
\tool treats port numbers up to 1024 individually as discrete values; beyond that it models ports in bins of size 100, as shown in  Figure~\ref{fig:ipport_arch} (b). In all, port numbers are represented 
as a 1670-categorical discrete attribute.
After being generated by \tool, if a port number is less than 1024 it corresponds to that particular port number, else
a port is sampled from a
uniform distribution of port numbers in the corresponding bin during post processing.

\subsection{Baselines}

We selected four different methods to serve as baselines for our method. This range for basic Gaussian Mixture Model, Bayesian Network to two recent deep learning approaches that use GANs
for synthetic data generation, which for brevity we refer to as GMM, BN, WPGAN, and CTGAN, respectively.
We compare \tool with these baselines and analyze the distribution factorization.

\textbf{Gaussian Mixture Model (GMM).} This assumes all attributes at a particular time step are independent of each other, and further that each row is independent. Thus it can be factorized as following:

\begin{subequations}
\begin{tabularx}{0.9\linewidth}{Xp{2cm}X}
  \begin{equation}
  \mathbb{P}(\textbf{x}) = \prod_{i=1}^{n} \mathbb{P}(\mathbf{x}_{i})
\label{jd:lv1}
  \end{equation}
  & &
  \begin{equation}
    \mathbb{P}(\textbf{x}_i) = \prod_{j=1}^{m} \mathbb{P}(\mathbf{x}_{i,j})
\label{jd:lv1-row}
  \end{equation}
\end{tabularx}
\end{subequations}

\textbf{Bayesian Network (BN).} As a traditional statistical approach, limited temporal or attributes dependence can be learnt based on the domain knowledge from experts.
For example, if $\textbf{x}_{i,j_1}$ is dependent on  $\textbf{x}_{i-1,j_1}$ and $\textbf{x}_{i,j_2}$, we can write it as a product of the conditional distributions (see Equation \ref{jd:lv2}).
The value $\mathbb{P}(\textbf{x}_{i,j_1} | \textbf{x}_{i-1,j_1}, \textbf{x}_{i,j_2})$ is the probability of the $j_1$ attributes of the $i$-th observation row, given the $(i-1)$-th $j_1$ attribute and the $i$-th $j_2$ attribute.
Considering the edge situation as well as utilizing the Bayes rule, we rewrite the distribution $\mathbb{P}(\textbf{x}_{i,j_1} | \textbf{x}_{i-1,j_1}, \textbf{x}_{i,j_2})$ as:

\begin{equation}
\label{jd:lv2}
\begin{split}
\mathbb{P}(\textbf{x}) &= \prod_{i=1}^{n} [\mathbb{P}(x_{i,j_1} | x_{i,j_2}, x_{i-1,j_1})\prod_{j=1, j \neq j_1}^{m}\mathbb{P}(x_{i,j})]\\
&= \mathbb{P}(x_1)\cdot \prod_{i=2}^{n} [\mathbb{P}(x_{i,j_1})\mathbb{P}(x_{i-1,j_1}|x_{i,j_1})\mathbb{P}(x_{i,j_2}|x_{i,j_2})]\\
& \quad \cdot \prod_{j=1, j \neq j_1}^{m}\mathbb{P}(x_{i,j})
\end{split}
\end{equation}

\textbf{WPGAN} \cite{ring2019flow} utilizes GAN to specifically generate network traffic flow data, while \textbf{CTGAN} \cite{xu2019modeling} utilizes GAN to generate general tabular data that contains both discrete and continuous attributes.
Both B3 and B4 assume attribute dependence at a certain time step but ignore temporal-wise dependence.
Thus the joint distribution can be factorized as Equation \ref{jd:lv1} only, while the factorization inside each row is untractable due to the GAN mechanism.

\subsection{Evaluation Metrics}
\label{subsec:metrics}
The evaluation of generative models is challenging
and subjective.
We use multiple metrics to compare them: likelihood, distribution comparison, domain knowledge rules test, and machine learning tasks performance comparison.

\textbf{Likelihood.} The likelihood function measures the goodness of a statistical model fitting a data sample.
However, the intrinsic difference between explicit density methods (GMM, BN, and \tool) and implicit density methods (WPGAN and CTGAN) makes it more challenging to compare them.
Goodfellow et al. \cite{goodfellow2014generative} states that there is no fair approach to directly compare the likelihoods of GAN models. Thus in this paper, we only compare the likelihoods between explicit density models, that is, GMM, BN and, \tool.

\textbf{Distribution and JS divergence} Although the goal of our work is to model joint distribution of a window of data, we also compare the marginal distributions of the individual attributes.
As a quantitative metric, we calculate Jensen-Shannon divergence between the distributions of the generated data $\textbf{D}_{synth}$ and the real data $\textbf{D}_{test}$ for each attribute.

\textbf{Domain knowledge test.} We use domain knowledge checks to evaluate the synthetic data quality. Since the application data set pertains to network traffic flow, we use several properties that such data needs to satisfy in order to be realistic \cite{ring2019flow}.

In addition to marginal distributions, we also explore network traffic specific distributions such as that of the number of unique destinations (in terms of IP addresses), and of number of bytes per IP address, and compare then with those distributions in real data. Similarly, we compare the top most frequently occurring port numbers.

\textbf{Machine learning application task.}
The final goal of generating synthetic data is to build machine learning models without using any real data. To evaluate whether the generated data is able to replace real data in a model training process, we 
select two tasks used for cybersecurity anomaly detection.
 One is a classification task while the other is regression; both use self-supervision. 

While these tasks may not seem to directly relate to cybersecurity, they are good at detecting anomalies, which may be caused, e.g., by a cybersecurity breach. The basic assumption is that different attributes in network traffic data have underlying dependencies, e,g., between the protocol field and other fields, during normal operation and these relationships may change in the presence of security anomalies. The models are built to capture this normal relationship, and then try to detect any changes in it during deployment.

The first task is predicting the transport protocol field in network traffic flow data, while the second task is to predict the number of bytes in the next network flow. In practice once trained these models are used for marking anomalies when the actual value significantly differs from the real one based on a hyperparameter threshold value.
We train a RandomForest model for the classification task, and a neural network model for the regression task.

As the evaluation metric, we use the F1 score, and the mean square error (MSE) for the two tasks, respectively. Since the classification task is an multi-class task, we apply the macro-F1 score which takes the average of all the category F1 scores. This ensures equal treatment of all classes even when the class distribution is skewed, as is likely for transport protocol where TCP dominates.
For both tasks we compare the cross-validation performance of the models trained on real and synthetically generated data.

%% file: evaluation.tex
\section{Experimental Results}

To demonstrate its effectiveness, we train and
evaluate \tool on a real network traffic data set.
However, to experiment with some architectural variations, we first use a simple simulated data set.

\subsection{Understanding STAN using Simulated Data}
\label{sec:simulated}

We built a simulated dataset with a simple random process whose dependence can be clearly controlled. We simulated a two-variable data distribution with the following formula and sampled 10,000 points data set $(X, Y)$ from it:
\[
x_t=0.9x_{t-1}+0.1\mathcal{N}
\]
and
\[
y_t=0.9x_{t}+0.1\mathcal{N}, 
\]
where $\mathcal{N}$ is standard normal distribution. We incorporate both temporal and attribute dependence with two attributes, $X$ and $Y$.

To train we apply a naive version of \tool, that passes  the input to mixture density neural layers directly, and GMM on the simulated data set. Then we generated data using both \toola and \toolb.

To measure how well dependence is captured we use
the Pearson correlation coefficient $R$ both for temporal dependence $R(X_t, X_{t-1})$ and attribute dependence $R(X_t, Y_t)$. In addition, we use two machine learning tasks to show that the synthetic data is able to serve as model training source to replace the original data.

\begin{figure}[ht]
\centering
\begin{subfigure}{.45\linewidth}
    \centering
    \includegraphics[width=\linewidth]{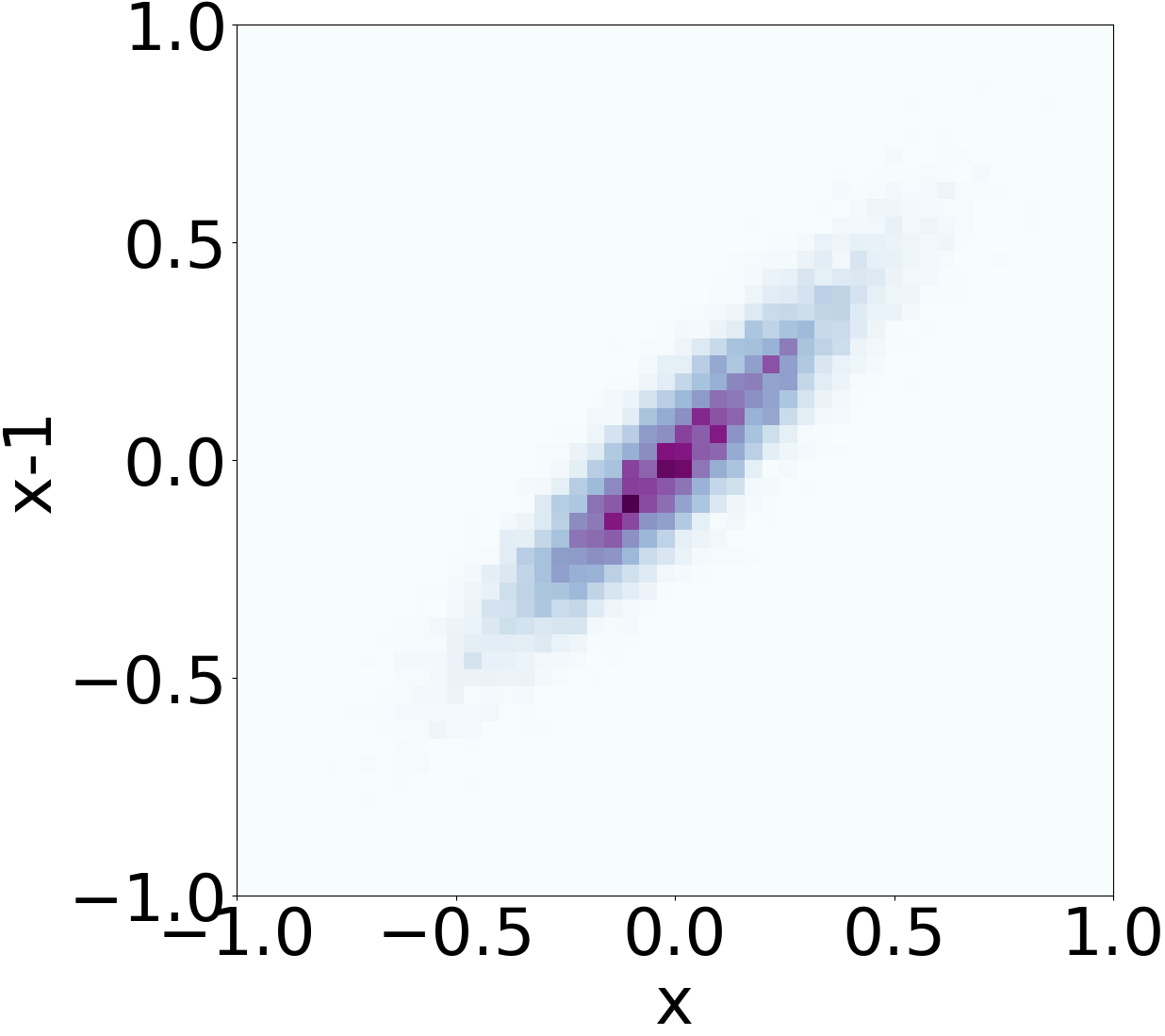}
    \subcaption{Original data,  $R(X_t, X_{t-1})$=$0.9$}
    \label{fig:simulated_xx_a}
\end{subfigure}%
\quad
\begin{subfigure}{.45\linewidth}
    \centering
    \includegraphics[width=\linewidth]{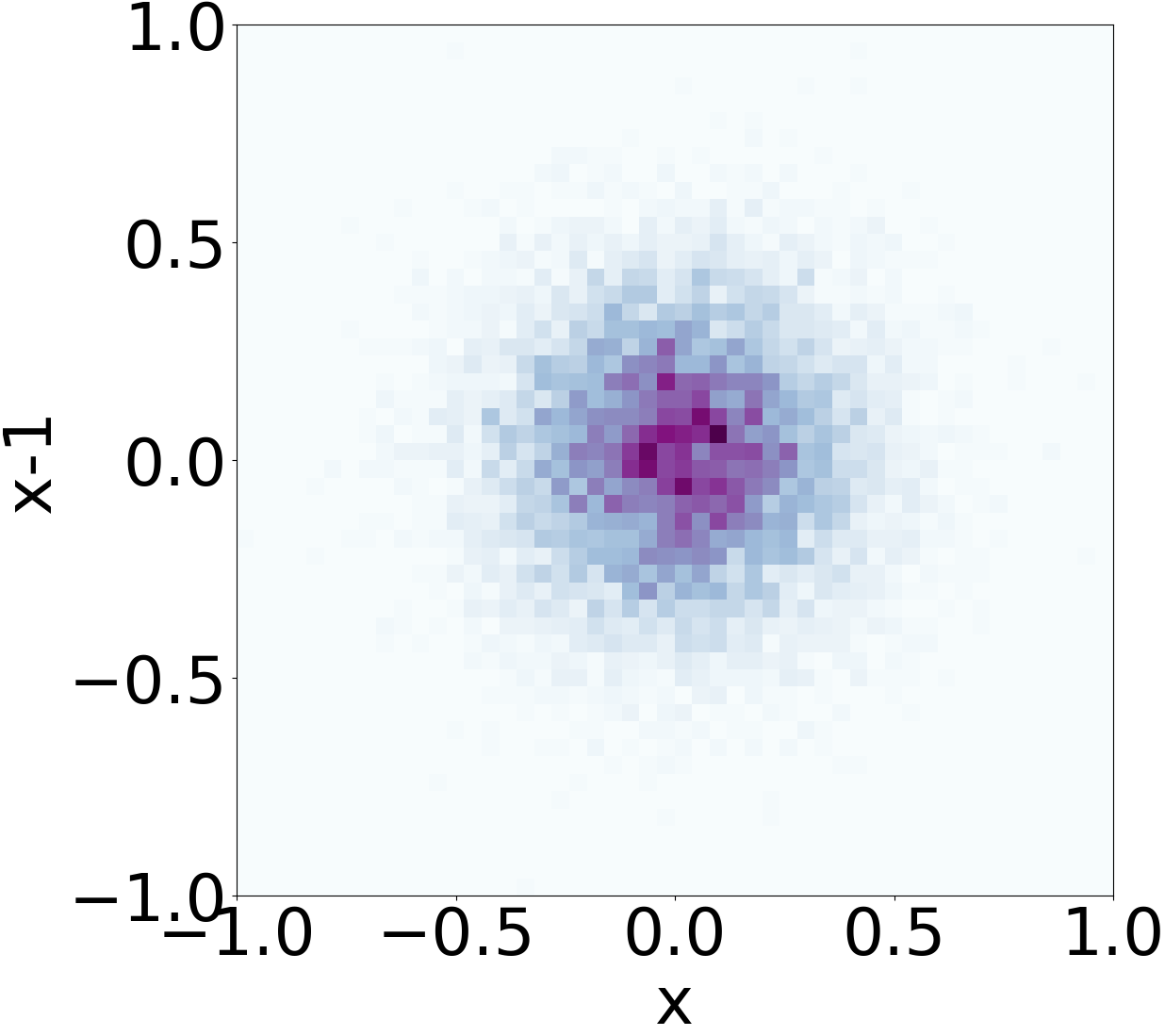}
    \subcaption{GMM synthesized data, $R(X_t, X_{t-1})$=$0$}
    \label{fig:simulated_xx_b}
\end{subfigure}
\quad
\begin{subfigure}{.45\linewidth}
    \centering
    \includegraphics[width=\linewidth]{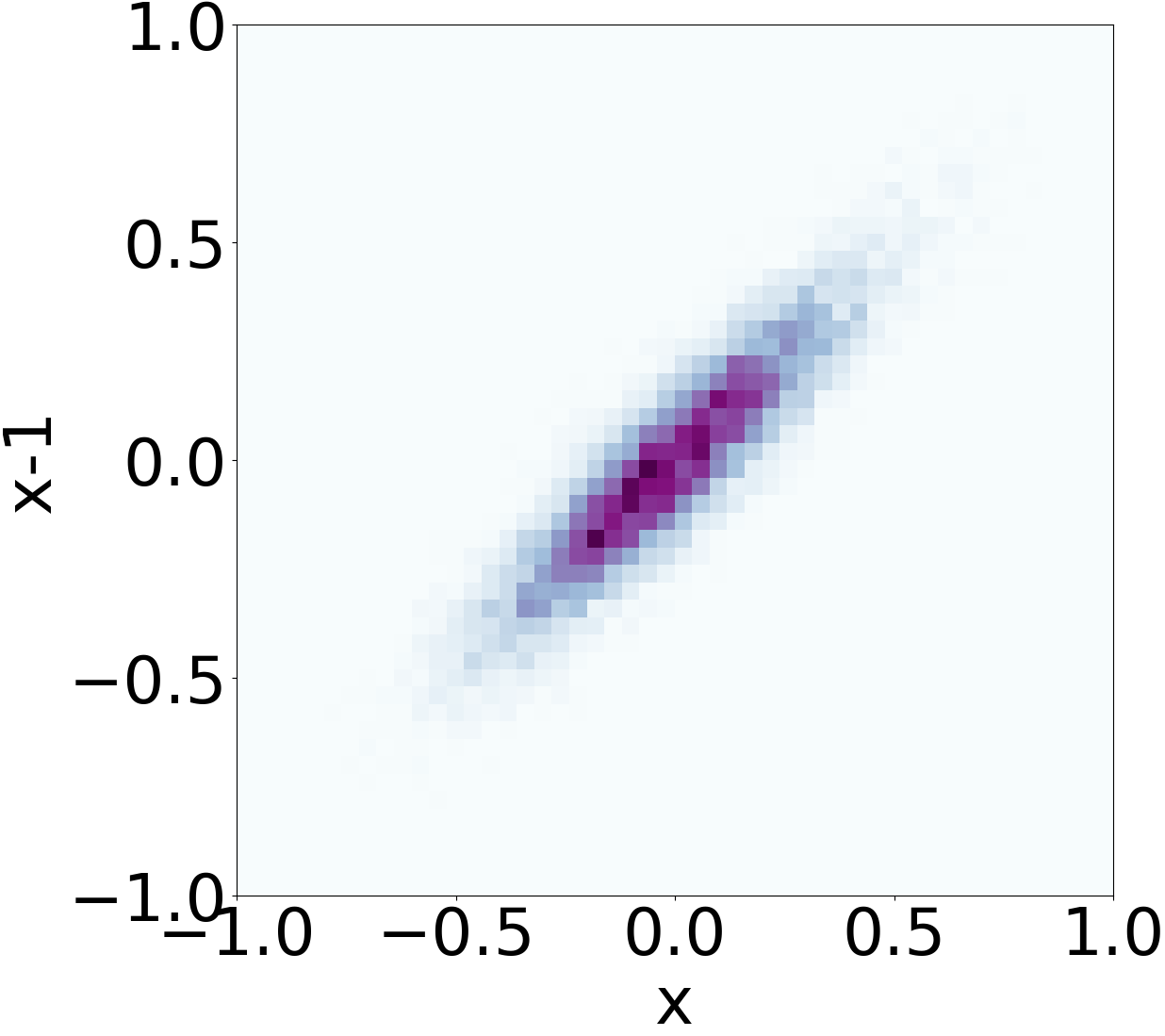}
    \subcaption{\tool mask A synthesized data, $R(X_t, X_{t-1})$=$0.9$}
    \label{fig:simulated_xx_c}
\end{subfigure}
\quad
\begin{subfigure}{.45\linewidth}
    \centering
    \includegraphics[width=\linewidth]{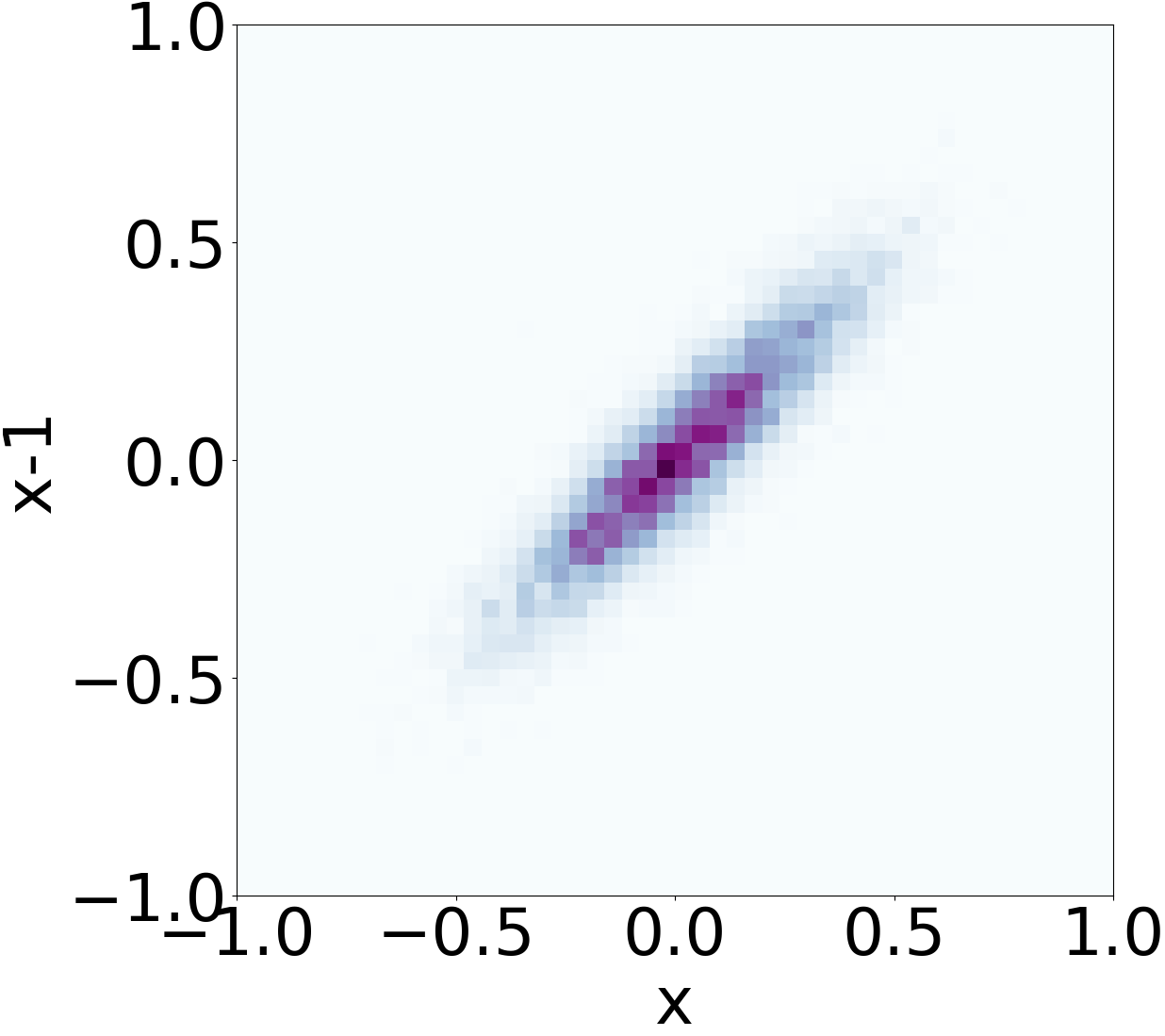}
    \subcaption{\tool mask B synthesized data, $R(X_t, X_{t-1})$=$0.9$}
    \label{fig:simulated_xx_d}
\end{subfigure}%
\caption[short]{Temporal dependence: $(X_t, X_{t-1})$ scatter plot of the simulated data and synthetic data with Correlation Coefficients $R$.}
\label{fig:simulated_xx}
\end{figure}

\begin{figure}[ht]
\centering
\begin{subfigure}{.45\linewidth}
    \centering
    \includegraphics[width=\linewidth]{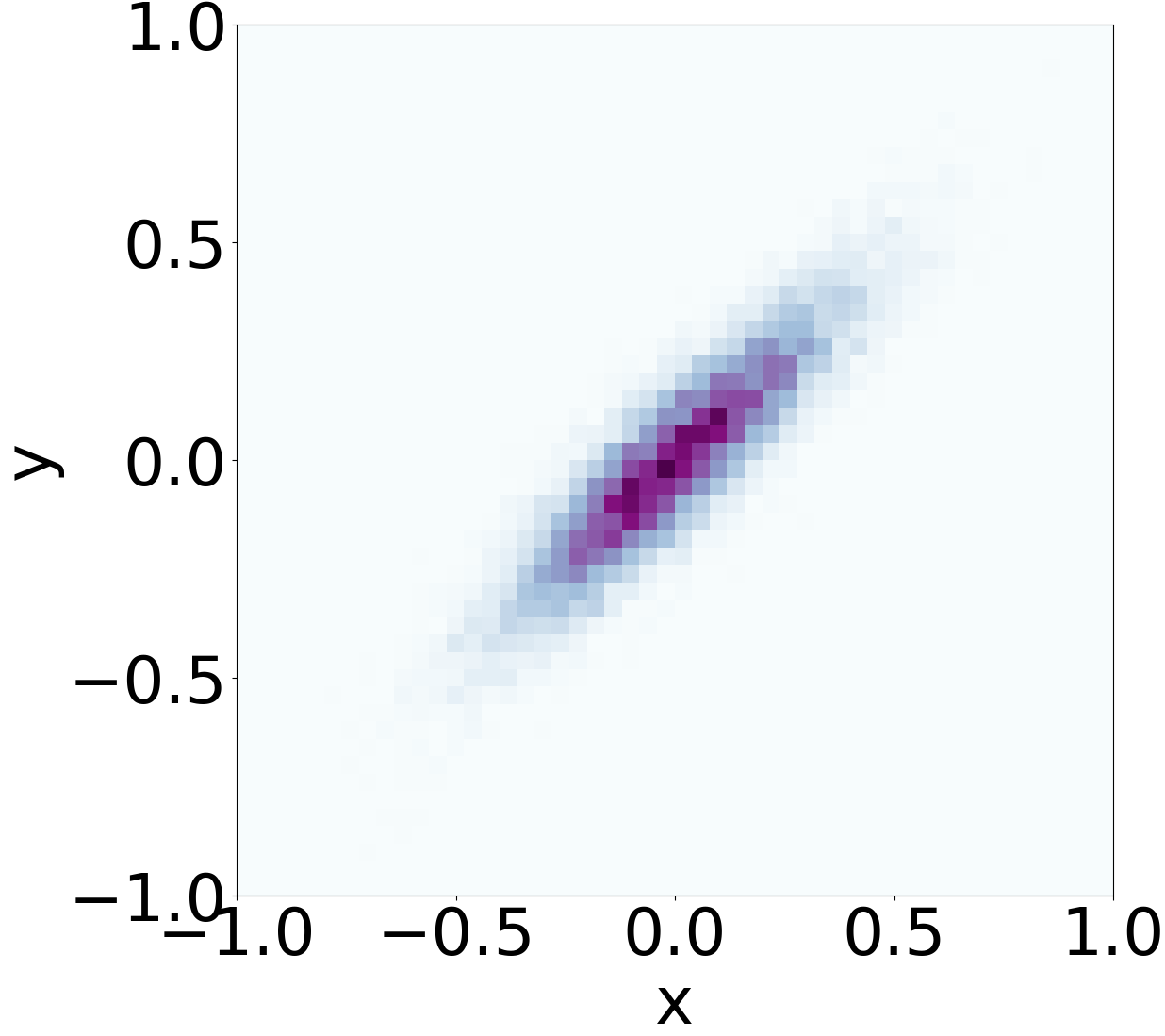}
    \subcaption{Original data,  $R(X_t, Y_t)$=$0.9$}
    \label{fig:simulated_xy_a}
\end{subfigure}%
\quad
\begin{subfigure}{.45\linewidth}
    \centering
    \includegraphics[width=\linewidth]{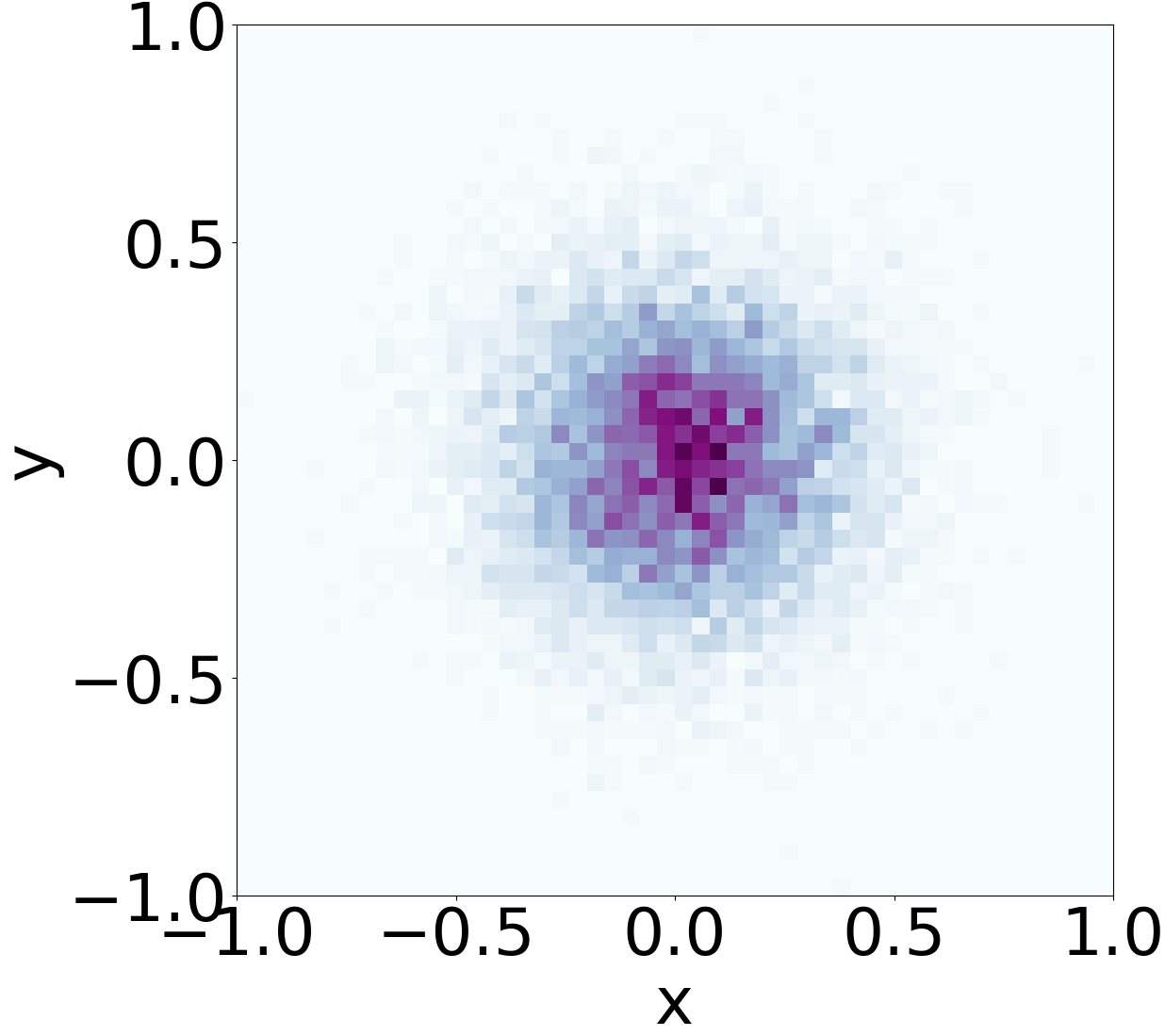}
    \subcaption{GMM synthesized data, $R(X_t, Y_t)$=$0$}
    \label{fig:simulated_xy_b}
\end{subfigure}
\quad
\begin{subfigure}{.45\linewidth}
    \centering
    \includegraphics[width=\linewidth]{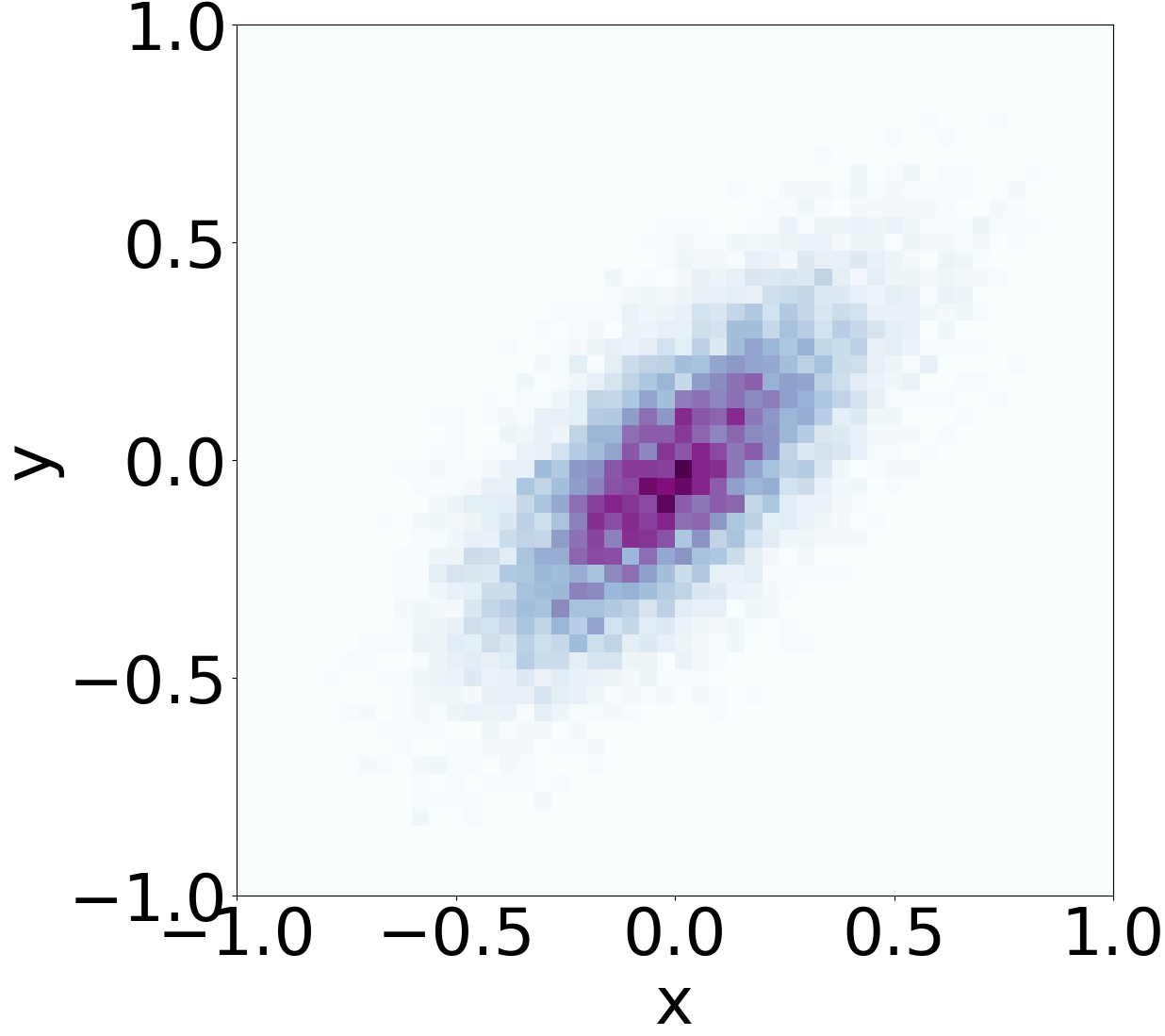}
    \subcaption{\tool mask A synthesized data, $R(X_t, Y_t)$=$0.7$}
    \label{fig:simulated_xy_c}
\end{subfigure}
\quad
\begin{subfigure}{.45\linewidth}
    \centering
    \includegraphics[width=\linewidth]{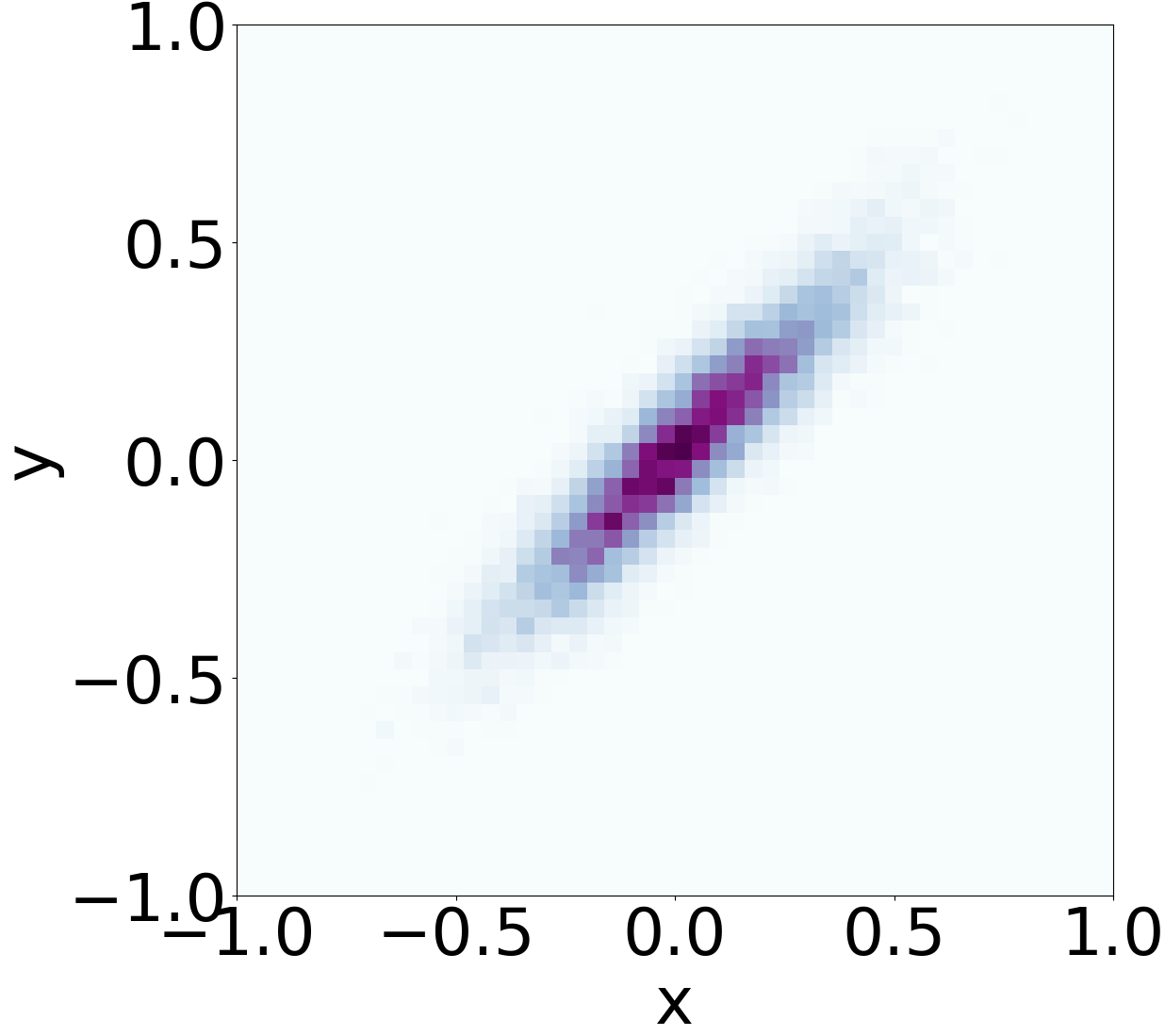}
    \subcaption{\tool mask B synthesized data, $R(X_t, Y_t)$=$0.9$}
    \label{fig:simulated_xy_d}
\end{subfigure}%
\caption[short]{Attribute dependence: $(X_t, Y_t)$ scatter plot of the simulated data and synthetic data with Correlation Coefficients $R$.}
\label{fig:simulated_xy}
\end{figure}

\textbf{Quantitative evaluation.}
We evaluated the correlation coefficient $R$ between both temporal dependence $R(X_t, X_{t-1})$ and attribute dependence $R(X_t, Y_t)$. Figure \ref{fig:simulated_xx} shows the scatter plots of $X_t$ and $X_{t-1}$  from four data sources (simulated data, GMM synthetic data, \tool with mask A synthetic data, and \tool with mask B synthetic data),
and Figure \ref{fig:simulated_xy} shows that for $X_t$ and $Y_t$.

Figures \ref{fig:simulated_xx_a} and \ref{fig:simulated_xy_a} show the dependence
in the original (simulated) data, across time and attribute, respectively, 
that the synthesizer needs to learn. The scatter plots 
$(x_{t-1}, x_t)$ and $(x_t, y_t)$  show strong linear relationship.
Figures \ref{fig:simulated_xx_b} and \ref{fig:simulated_xy_b} show that independently learned marginal distribution is unable to generate data with 
temporal and attribute-wise dependence. Figure \ref{fig:simulated_xx_c} and \ref{fig:simulated_xx_d} show \toola and \toolb perform similarly 
and are able to generate data with the same temporal dependence as in the simulated (original) data. However, Figure \ref{fig:simulated_xy_c} and \ref{fig:simulated_xy_d} show that \toolb, when explicitly conditioned on the same-row attribute context in its convolution perceptive field, generates better attribute-wise dependence than \toola.

Since mask A and mask B represent \textit{conditional independence} and \textit{explicit dependence} respectively, we summarize through this observations:
\begin{itemize}
\item Both conditional independence and explicit dependence provide reasonable approximation for temporal dependence. The $R(X_t, X_{t-1})$ of simulated data, synthetic data generated using \tool mask A, and synthetic data generated using \tool mask B are all same (0.9).
\item Conditional independence provides a reasonable same-row attribute approximation, while explicit dependence performs better. The $R(X_t, Y_t)$ of simulated data, and synthetic data generated using \tool mask B is 0.9; while that of synthetic data generated using \tool mask A is 0.7.
\end{itemize}

\textbf{Machine learning tasks.}
We also used the simulated data and the corresponding synthetic data for training two 
machine learning tasks (using the scikit-learn Python library). Then we evaluate the performance (Mean Square Error) of the trained machine learning models on the simulated test data. Simulated training data and simulated test data follow the same distribution.

\begin{itemize}
\item \textbf{T1}: predict  $y_t$ given $x_t$ (row attribute dependence).
\item \textbf{T2}: predict  $x_{t+1}$ given $x_t$ (temporal dependence).
\end{itemize}

Table \ref{table:simulated_task} shows that a machine learning model trained only on synthetic data generated by \tool produces similar test loss as that trained on the original 
simulated test data.

\begin{table}[ht]
\centering
\begin{tabular}{l|l|l}
\toprule
Training Data       & MSE(\textbf{T1})      & MSE(\textbf{T2}) \\ 
\midrule
Simulated data & 0.010 & 0.01\\
GMM & 0.050 & 0.05 \\ 
\midrule
STAN mask A & 0.013       &  0.01\\ 
STAN mask B & 0.010 &  0.01\\ 
\bottomrule
\end{tabular}
\caption{Mean Square Error of the two tasks}
\label{table:simulated_task}
\end{table}

\subsection{Real network traffic data}
\label{sec:real_eval}

\textbf{Data set.}
Network traffic data is typically a multivariate time-series. 
 A common format is called {\em netflow}, where each row represents a unidirectional network traffic connection or flow. We selected a \textit{netflow} data set for our experiments since it 
 is a good representative format for network traffic data in general.
 Here we use a large publicly available {\em netflow} data set. Typically each row consists of the following attributes: timestamp at the end of a flow (te), duration of flow (td), packets exchanged in the flow (pkt), and the corresponding number of bytes (byt), source IP address\footnote{We only consider IPv4 addresses here.} (sa), destination IP address (da), source port (sp), destination port (dp), flags (flg), and transport protocol (pr). Each row $x_i$ can be expressed as a tuple of ($te_i, byt_i, sa_i, da_i, pr_i$, etc). Table \ref{table:traffic} shows typical attributes, their types and example values.

\begin{table}[ht]
\centering
\begin{tabular}{l|l|l}
\toprule
Attribute       & Type      & Example \\ 
\midrule
timestamp & continuous & 2016-04-11 00:02:15 \\ 
duration        & continuous & 0.344 \\ 
transport protocol & categorical & TCP \\ 
source IP address & categorical & 85.201.196.53 \\ 
source port & categorical & 19925 \\ 
dest. IP address & categorical & 42.219.145.151 \\ 
dest. port & categorical & 80 \\ 
bytes & numeric & 11238 \\ 
packets & numeric & 11 \\
TCP flags & categorical & .A..SF \\
\bottomrule
\end{tabular}
\caption{Overview of typical attributes in flow-based data.}
\label{table:traffic}
\end{table}

We apply \tool on a publicly available benchmark \textit{netflow} data set, UGR'16 \cite{macia2018ugr}, which contains
large scale traffic data captured by a Tier-3 ISP cloud service provider. 
First, we selected a week of data (April week3) data to focus on.
We selected this week this week since it looked interesting in terms of volume of traffic and the number of events marked. 
Second, we
randomly selected flows related to 90 users (essentially IP addresses) based on the number of traffic flows per user distribution. 
Third, we extracted one day's (Monday) data to be the $\textbf{D}_{historical}$ and another day (Tuesday) of the same user group and the same week to be the $\textbf{D}_{test}$.
Following this strategy, we selected 1,531,126 samples for the $\textbf{D}_{historical}$ and 1,952,702 samples for the $\textbf{D}_{test}$.
Ten percent of $\textbf{D}_{historical}$ are selected out to serve as training validation data $\textbf{D}_{validation}$.

\textbf{\textit{netflow} data processing.}
To ensure the trained model is a practical and robust tool to synthesize network traffic flow data, we normalize the raw \textit{netflow} data for ease of processing by the neural network. Also, the neural network predicted values are transformed back into the original scale. 

The inputs to the neural model are pre-processed to facilitate training. The numerical attributes are min-max scaled; for the categorical attributes, we apply one-hot encoding. Specifically, for the protocol attribute we use a three-way softmax (for TCP, UDP and other). Note that for simplicity we consider only three protocol categories since TCP and UDP are the most prevalent; it would be easy to extend it to more categories if needed.  
For source and destination port number attributes, we handle well-known and other ports differently as described in Section \ref{sec:model_ipport}.
Instead of modeling timestamps of individual flows, we model the time deltas between them.

\textbf{Training hyperparameters.}
Our models are trained on four Tesla P100 GPUs using the Pytorch toolbox. From the different parameter update rules tried, the Adam~\cite{kingma2014adam} algorithm
gives best convergence performance and is used for all experiments.
The learning rate schedules were manually set to the highest values that allowed fast convergence: 0.001 for mixture density neural layers and 0.01 for softmax layers. The batch sizes are also manually set for the experiments. For UGR16, we use as large a batch size as that showed 
quick converge; this corresponds to 512 time windows input per
batch. We use pre-processing to prepare data batches that can be trained in parallel and accelerate the training and generation process.
For mixture density neural layers, we select 10 as the Gaussian components to be learned based on the cross-validation.
For the initial convolution network layer parameters, we sample from a Uniform distribution whose boundary is the standard deviation of the kernel size, i.e. [$-\frac{1}{3*3}$, $\frac{1}{3*3}$].

\textbf{Likelihood.}
\label{subsec:eval_likelihood}
For each data point (each row), we can directly calculate the row likelihood by factorization equations.
In our case, explicit density generative models (GMM, BN and \tool) clearly define the distribution for each attributes and for those we can evaluate the modeled distribution directly via individual attribute distributions.
For simplicity, we discretize
continuous variables to validate their negative log-likelihood value for all the baselines and attributes, based on the variable value range and the data set size.
In Table \ref{table:indiv_nll} we report the negative log likelihood (NLL) of a few attributes as modelled by \tool and baselines GMM and BN. Note that we are unable to generate IP addresses and port numbers using GMM and BN, so the NLL for those attributes is not compared in the table. \tool produces better results for both continuous and discrete attributes.

\begin{table}[ht]
\centering
\begin{tabular}{l|l|l|l|l}
\toprule
Model & bytes & packet & time duration & transport protocol \\ \midrule
GMM & 4.85 & 3.78 & 1.81 & 0.341 \\ 
BN & 3.90 & 2.62 & 0.97 & 0.344 \\ 
\midrule 
\tool & 2.34 & 1.73 & 0.59 & 0.002 \\
\bottomrule
\end{tabular}
\caption{Attribute negative log likelihood of models evaluated on $\textbf{D}_{validation}$ (lower is better).}
\label{table:indiv_nll}
\end{table}

\textbf{Data Synthesis.}
Once a \tool model is trained on $\textbf{D}_{historical}$, it is able to sequentially generate any length of {\it netflow} data $\textbf{D}_{synth}$. As described in Section \ref{sec:method}, \tool starts the generation process by sampling from the trained marginal distribution without any input requirement, and then autoregressively generates row by row conditioned on the prior rows context. To fairly compare to $\textbf{D}_{test}$, we used \tool to generate $\textbf{D}_{synth}$ that includes 1,208,182 samples for the same set of $\textbf{D}_{historical}$ users.
Note that since we are generating data in one day's range (based on the generated delta time attribute $dt$ and the accumulated timestamp), the total number of samples is not directly determined by any hyperparameter. In the rest of this section, we evaluate the comparability between the $\textbf{D}_{synth}$ and $\textbf{D}_{test}$.

\begin{figure}[ht]
\centering
\includegraphics[width=\linewidth]{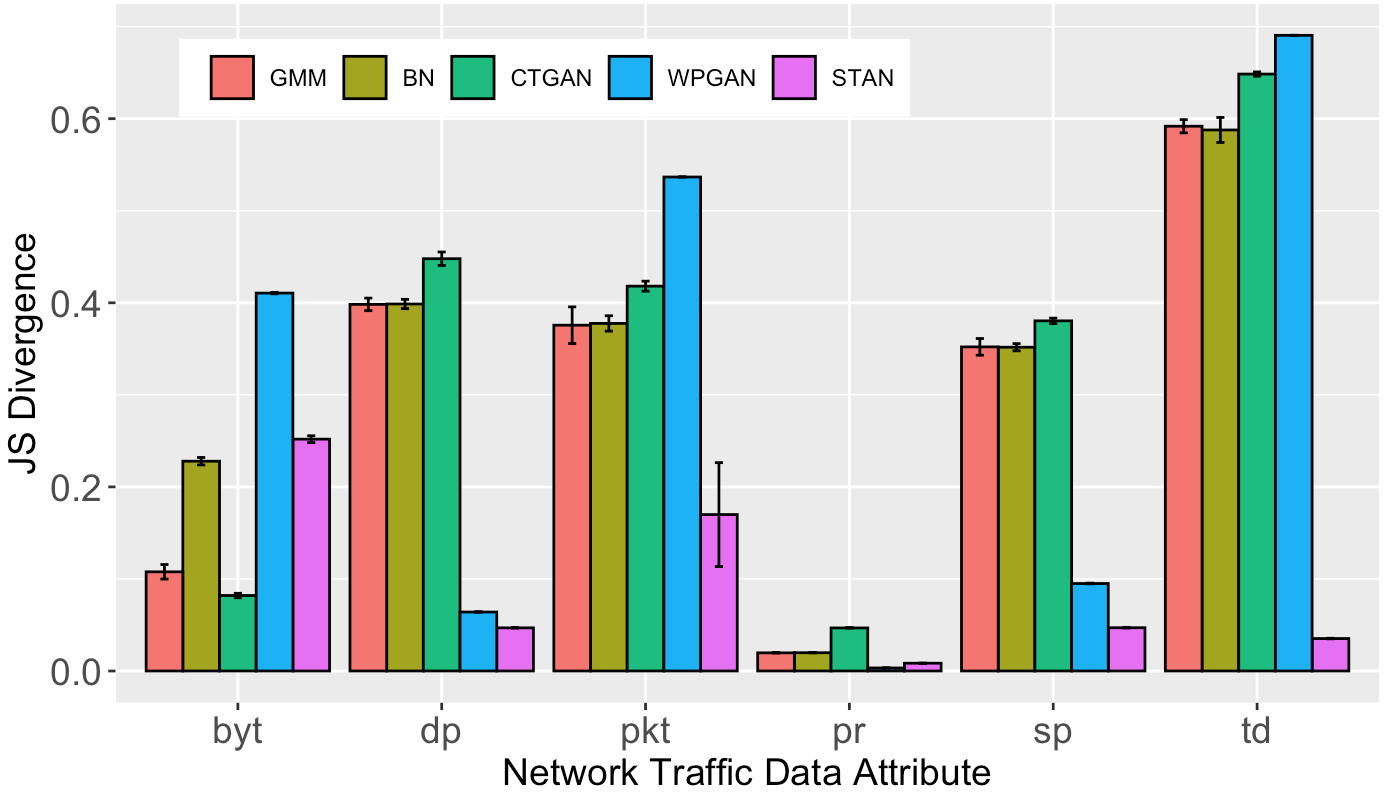}
\caption{JS divergence between attribute marginal distribution between $\textbf{D}_{test}$ and $\textbf{D}_{synth}$ from \tool as well as that from baselines.}
\label{fig:js_div}
\end{figure}

\textbf{Distribution and JS divergence.}
Figure \ref{fig:js_div} shows the individual JS divergence of the marginal distribution of both the continuous and discrete attributes. \tool captures the marginal distribution well for most attributes.
Even though GMM precisely models
the marginal distribution of the training data set, it does not perform as well as \tool on the test data set. We believe this is because the marginal distribution over days is non-stationary. 

\textbf{Observation 1:} \emph{
\tool models the marginal distribution better than baseline GMM. 
}

\textbf{Domain knowledge test.}
As described earlier we perform basic sanity checks 
specified as rules on that need to be satisfied by generated flow-based network data. There are two basic types of rules -- those that apply to an individual attribute and those that check relationships between multiple attributes. Note that individual tests on port numbers and protocols are not needed since they are modeled as categorical variables. 
We highlight five tests here which are summarized in Table \ref{table:dkcheck}.  
\tool performs well in all three.
\begin{itemize}
  \item Test 1: Validity of IP address. Source IP address should not be multicast (from 224.0.0.0 to 239.255.255.255) or broadcast (255.xxx.xxx.xxx); Destination IP address should not be of the form 0.xxx.xxx.xxx. (Note that source address can be all zeros, e.g. for DHCP requests.)

  \item Test 2: Number of bytes/packets minimum size: The minimum value for $pkt$ attribute is 1 and the minimum value for $byt$ attribute depends on the transport protocol. For a TCP flow packet, the minimum size is 40 bytes (20 bytes for IP header + 20 bytes for TCP header); similarly for a UDP flow packet, the minimum size is 28 bytes (20 bytes for IP header + 8 bytes for UDP header).
  
  \item Test 3: Relationship between number of bytes (byt) and number of packets (pkt). Based on the protocol the following relationship exists between these attributes.
  For a TCP flow,
  \[
    40*pkt \leq byt \leq 65535*pkt
  \]
  Similarly, for a UDP flow,
  \[
    28*pkt \leq byt \leq 65535*pkt
  \]  
   The maximum packet size for both TCP (assuming maximum MTU) and UDP is 64K bytes.
    
  \item Test 4: If the transport protocol is not TCP, then the flow should not have any TCP flags.
  
  \item Test 5: If the flow describes normal user behavior and the source port or destination port is 80 (HTTP) or 443 (HTTPS), the transport protocol must be TCP\footnote{There is an effort to move HTTP/S to UDP, referred to as QUIC\cite{quic}. We use this test here since there was no QUIC traffic in our netflow dataset. However, this test may not be relevant for a different dataset.}.
  
\end{itemize}

Domain specific checks are performed to verify semantic consistency in the generated data. Depending on the data set, additional such checks can be added. If a generated data set performs poorly on these tests, modeling of different attributes in the traffic data set such as IP addresses and port numbers can be modified to capture the required semantic constructs, including possibly using distinct bins for modeling specific port numbers above 1024 as is now done for port numbers less than 1024.

\begin{table}[ht]
\centering
\begin{tabular}{l | l| l| l|l|l}
\toprule
           & Test 1 & Test 2 & Test 3 & Test 4 & Test 5 \\
\midrule
Real Data  & 99    & 100     & 100   & 100  & 100  \\
\midrule
GMM & 98    & 66     & 48 & -- & 77    \\
BN & 98    & 67     & 49 & -- & 78    \\
WPGAN \cite{ring2019flow}     & 99     & 75     & 72  & 97  & 99  \\
CTGAN \cite{xu2019modeling}    & 99     & 93     & 74   & -- & 99   \\
\midrule
\tool      & \textbf{99}    & \textbf{93}     & \textbf{93}   & \textbf{100} & \textbf{99}     \\
\bottomrule
\end{tabular}
\caption{Passing percentage of domain knowledge tests. Dash (-) means the methods are not able to generate `flags' attribute.}
\label{table:dkcheck}
\end{table}

\begin{figure}[ht]
\centering
\begin{subfigure}{.52\linewidth}
    \centering
    \includegraphics[width=\linewidth]{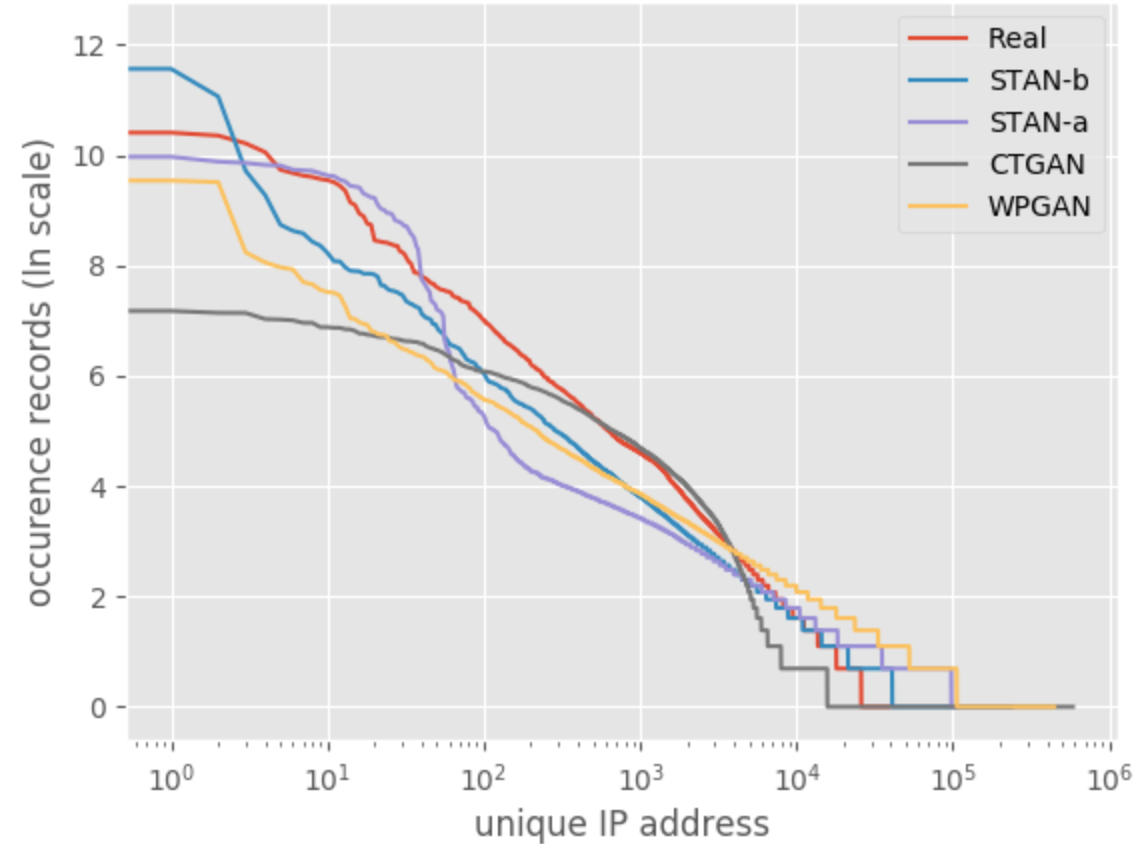}
    \subcaption{Unique IP address occurrence distribution. The y-axis (ln-scale) is the number of distinct IP addresses contacted.}
    \label{fig:ip_power}
\end{subfigure}
\begin{subfigure}{.52\linewidth}
    \centering
    \includegraphics[width=\linewidth]{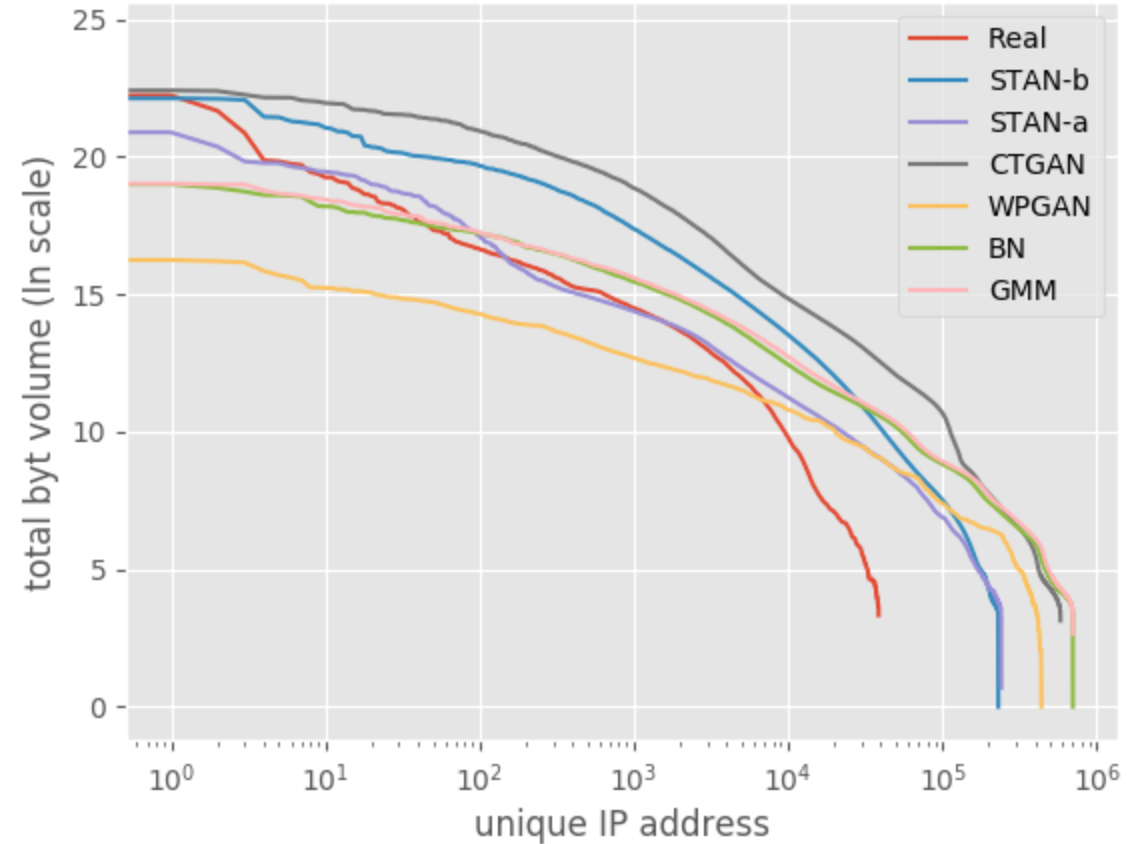}
    \subcaption{Unique IP volume ($bytes$) distribution. The y-axis (ln-scale) is the total traffic in bytes related to a user (unique IP address).}
    \label{fig:ip_vol}
\end{subfigure}

\caption[short]{IP address characteristics. The x-axis (log-scale) represents unique user (IP address) that occurs in the \textit{netflow} traffic.}
\label{fig:ip_characteristic}

\end{figure}

\textbf{Marginal distributions.}
We explore marginal distributions of \textit{netflow} attributes in the generated data. 
Figure \ref{fig:ip_power} shows the distribution of number of unique IP addresses each user communicates with. Typically, such distributions  follow a power law distribution \cite{METCALF201623}. \tool generated synthetic data comes closest to the real network data, including the unique IP address maximum occurrence value, the total number of different unique IP addresses, and the distribution curve. Meanwhile, Figure \ref{fig:ip_vol} shows the distribution of the total number of bytes exchanged
related to each user (IP address). Again, \tool synthetic data follows the real data distribution closer than the baseline methods.
It is worth noting that \tool learned these distributions without any explicit design choices on our part, e.g., in the loss function. It correctly inferred by marginal distribution by virtue of learning the joint distribution.

\begin{figure}[ht]
\centering
\begin{subfigure}{.59\linewidth}
    \centering
    \includegraphics[width=\linewidth]{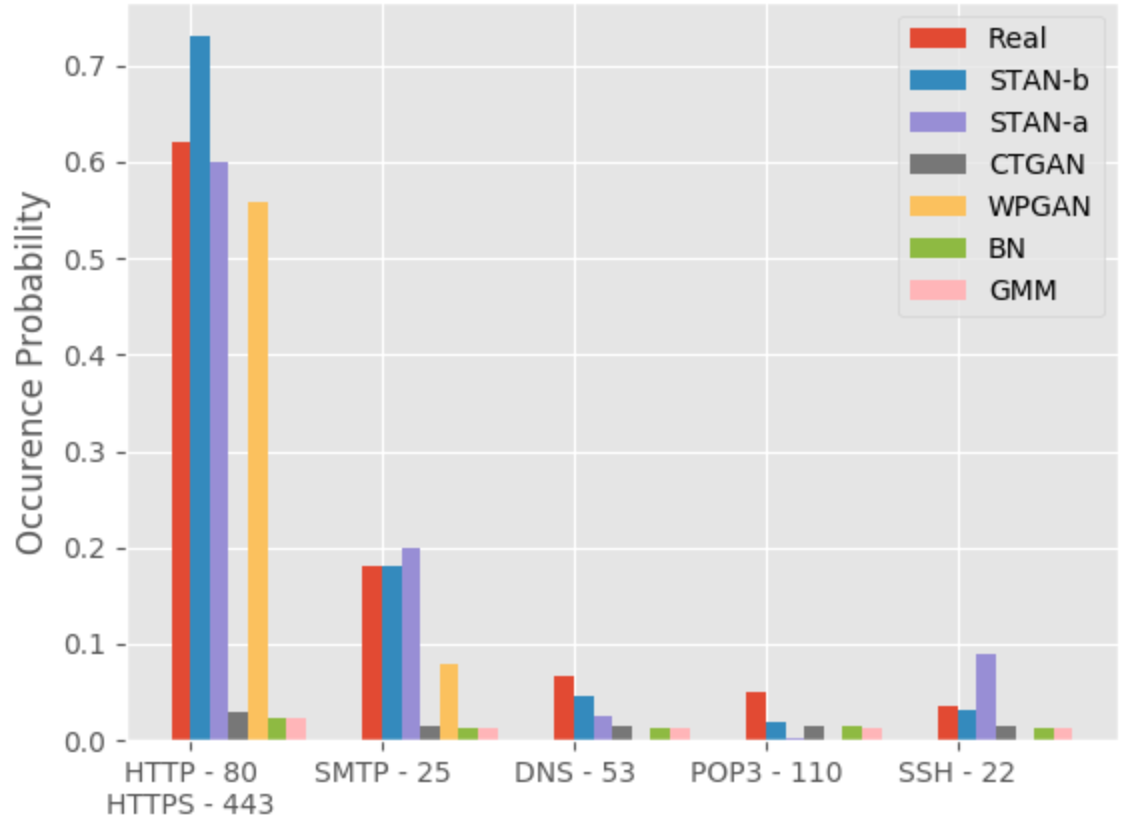}
    \subcaption{Top 5 most frequently occurring TCP ports and their related service}
    \label{fig:tcp_top5}
\end{subfigure}
\begin{subfigure}{.59\linewidth}
    \centering
    \includegraphics[width=\linewidth]{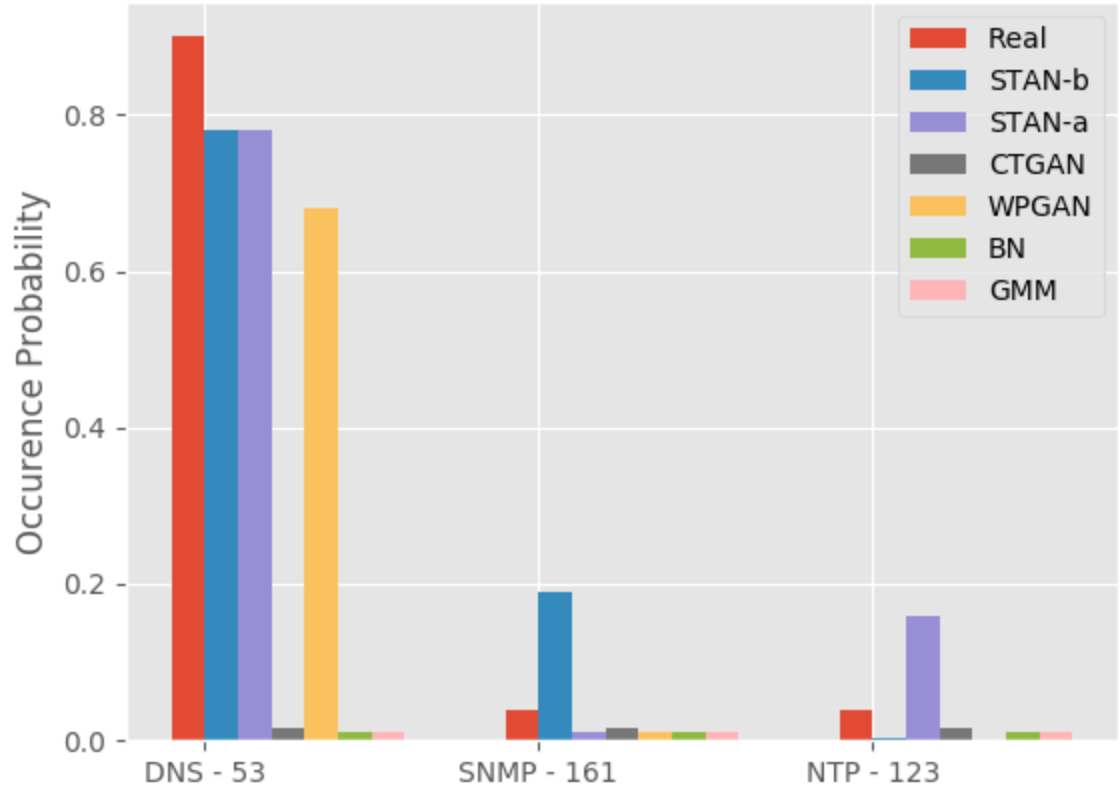}
    \subcaption{Top 3 most frequently occurring UDP ports and their related service}
    \label{fig:udp_top3}
\end{subfigure}

\caption[short]{Port number characteristics.}
\label{fig:port}

\end{figure}

We also compare port number distribution between real and synthetic data. 
Based on the real data, we select the top 5 TCP services and top 3 UDP services, which appear most frequently and the occurrence ratio are greater than 1\% in the entire TCP or UDP traffic. Figure \ref{fig:tcp_top5} shows the occurrence probability of that service in the entire TCP traffic records. We have 80/443 port for HTTP/HTTPS service (Hypertext Transfer Protocol / Hypertext Transfer Protocol Secure), 25 port for SMTP (Simple Mail Transfer Protocol), 53 port for DNS (Domain Name System), 110 port for POP3 (Post Office Protocol, version 3), and 22 port for SSH (Secure Shell). Similarly in Figure \ref{fig:udp_top3}, we have 53 port for DNS service , 161 port for SNMP (Simple Network Management Protocol), and 123 port for NTP (Network Time Protocol). We find \tool performs well at generating a port number distribution similar to real data. Further, this implies that \tool does a good job of capturing application level traffic, which can be mapped to different ports. 
While WPGAN with IP2Vec component is the second best, the rest of baselines (GMM, BN, CTGAN) perform poorly.

\begin{figure}[ht]
\centering
\begin{subfigure}{.59\linewidth}
    \centering
    \includegraphics[width=\linewidth]{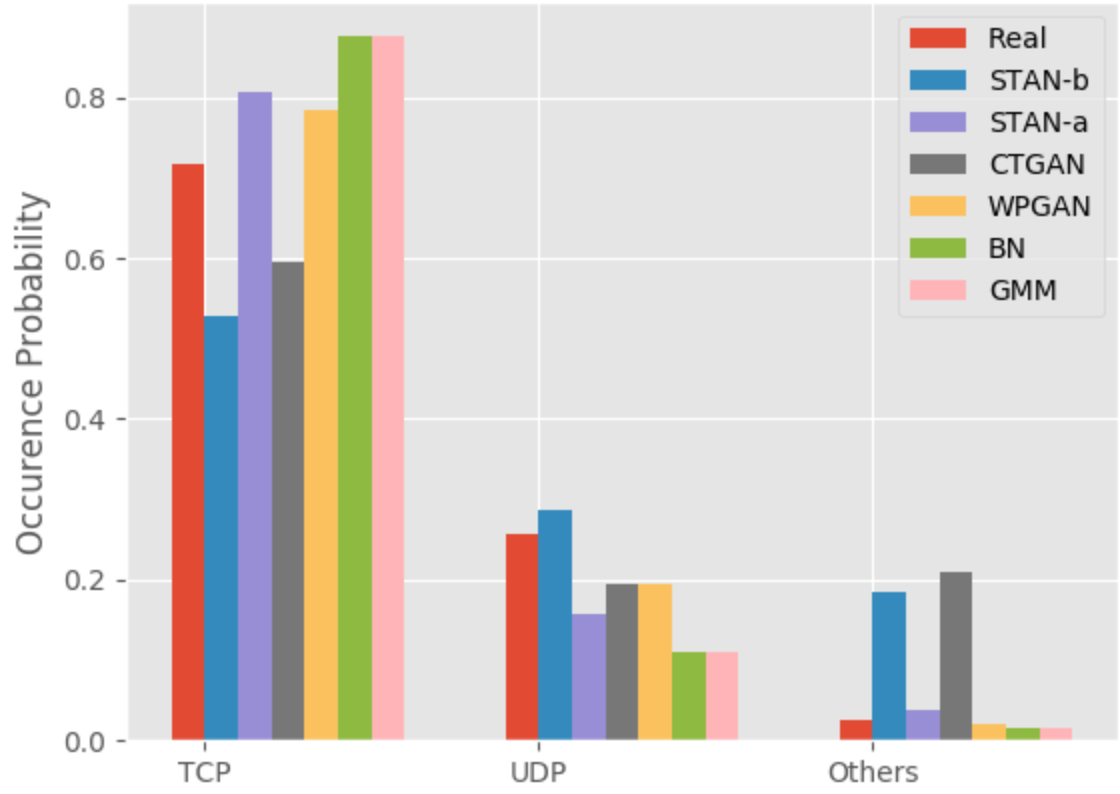}
    \subcaption{Marginal protocol distribution}
    \label{fig:pr_distribution}
\end{subfigure}
\begin{subfigure}{.59\linewidth}
    \centering
    \includegraphics[width=\linewidth]{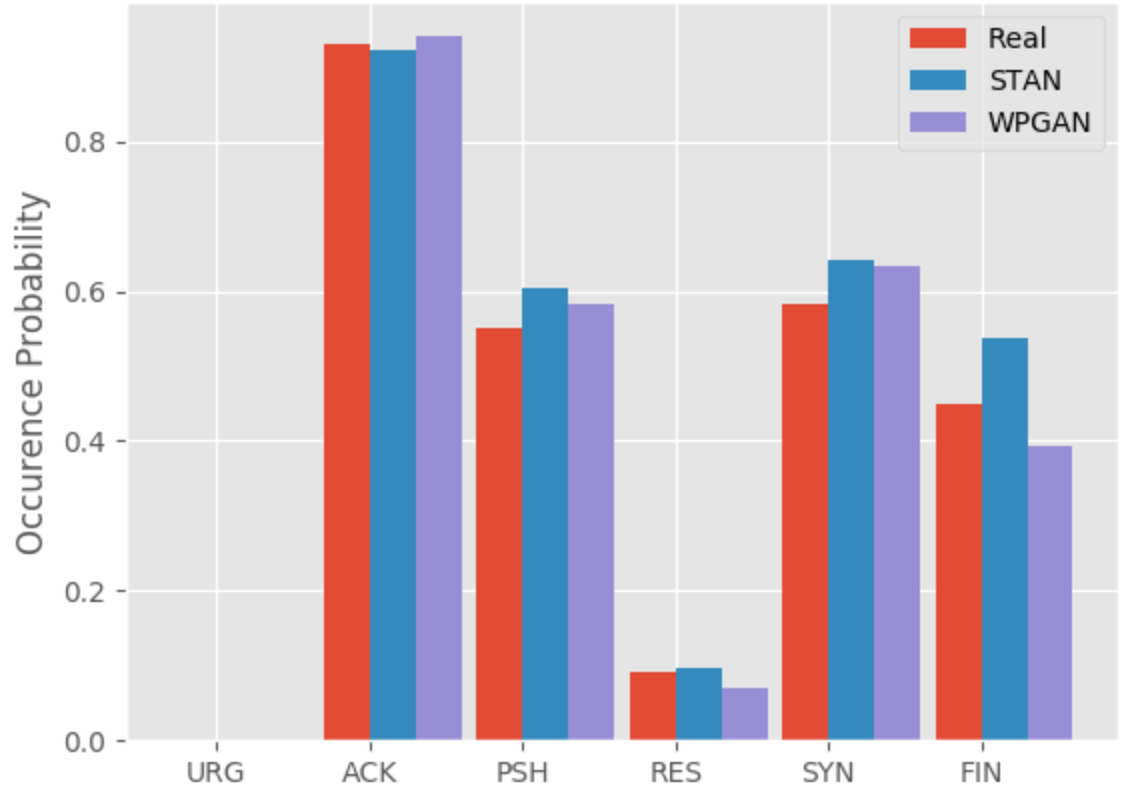}
    \subcaption{TCP flag distribution, includings six flags: URG, ACK, PSH, RES, SYN, and FIN}
    \label{fig:flag_distribution}
\end{subfigure}

\caption[short]{Transport protocol and TCP flags characteristics.}
\label{fig:port_pr_flg}

\end{figure}

Figures \ref{fig:pr_distribution} and \ref{fig:flag_distribution} show the marginal distribution of protocol and flags attributes respectively. As expected, TCP is the dominant protocol followed by UDP. For simplicity, we group the other protocols as `other'. For both protocol and flags, \tool generated data shows a similar distribution to real data. 

\textbf{Observation 2:} \emph{
Compared to baselines, \tool can learn the IP and port characteristics better without using domain knowledge or other design tuning.
}

\textbf{Cybersecurity application tasks.}
Finally, we test our synthetic data on two cybersecurity machine learning applications, to detect anomalies using self-supervision.
One of the tasks is a classification problem, and the other is a regression problem. The goal is to figure out
whether it is possible to fully substitute real data with synthetic data for training machine learning models.

A series of models are trained on real test data. We start our training from using a complete $\textbf{D}_{test}$ (real data) and successively decrease the amount of real data until no data from $\textbf{D}_{test}$ is used.
Another series of models are trained similarly using the real test data; however, instead of simply removing certain amount of data from $\textbf{D}_{test}$, we substitute the indicated amount of data with our synthetic data $\textbf{D}_{synth}$, so that the total amount of data is kept unchanged.

In the following two tasks, we use $\textbf{D}_{test}$, which is unseen and never used in the synthesizer training process.
For the synthetic data $\textbf{D}_{synth}$, every synthesizer model generates five sets of synthetic data sample, so  we can compute error bars. Five-fold cross validation is used to get a robust estimate of the measurements.

\begin{figure}[ht]
\centering
\begin{subfigure}{.75\linewidth}
    \centering
    \includegraphics[width=\linewidth]{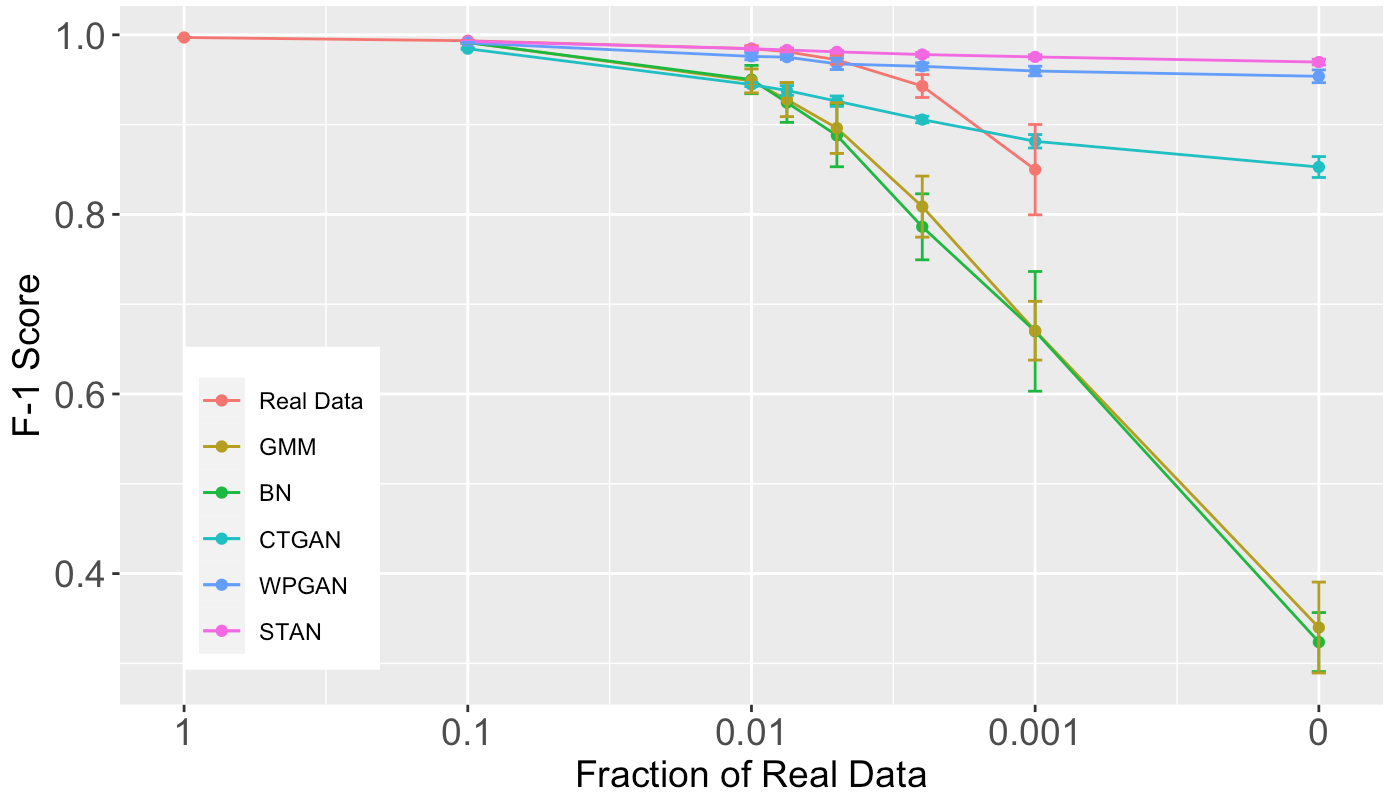}
    \subcaption{F1-score of same-row Protocol prediction Task}
    \label{fig:task1}
\end{subfigure}
\\
\begin{subfigure}{.75\linewidth}
    \centering
    \includegraphics[width=\linewidth]{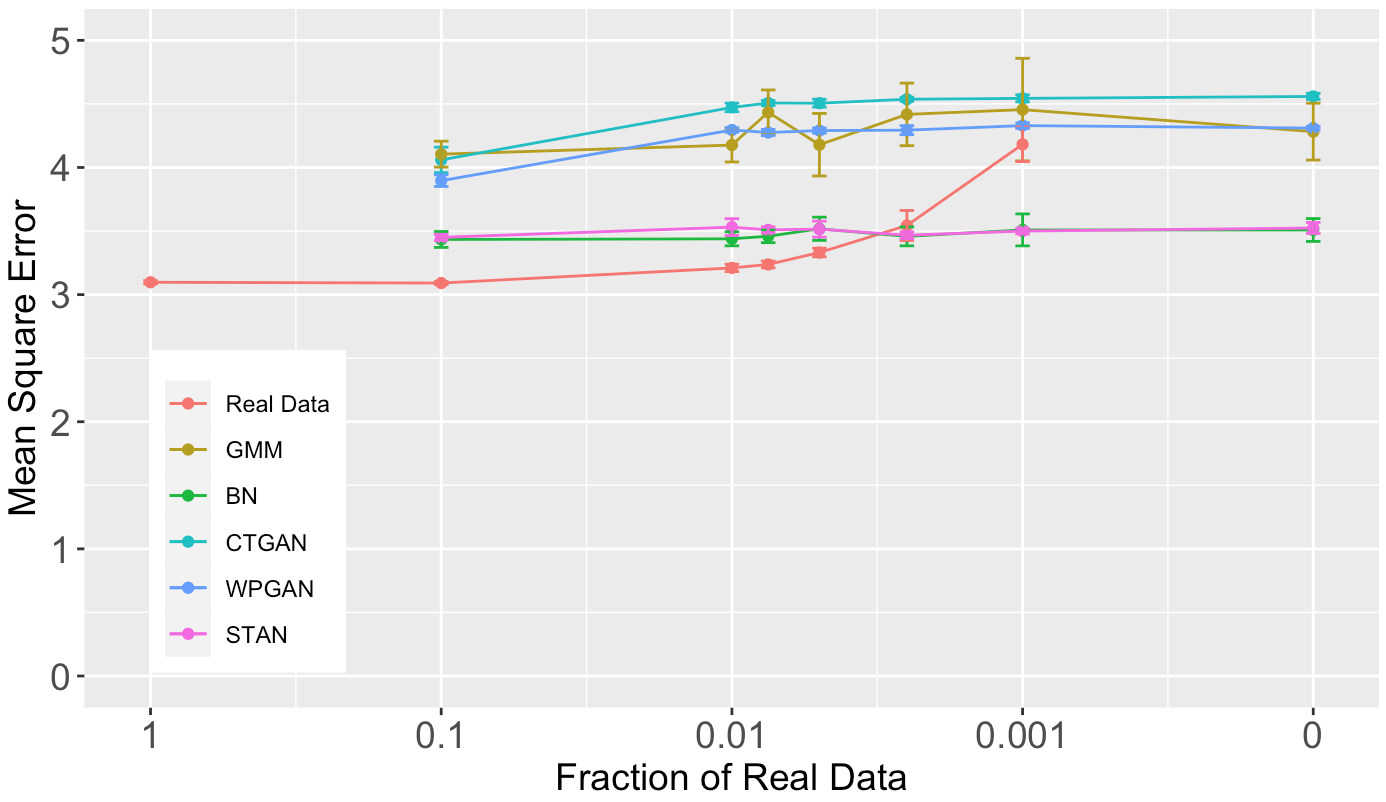}
    \subcaption{Mean Square Error of $bytes$ Value Forecasting Task}
    \label{fig:task2}
\end{subfigure}
\caption[short]{Real application task performance.}
\label{fig:task_results}

\end{figure}

\textbf{Task1: protocol forecasting.} Fig. \ref{fig:task1} shows 
the F-1 scores achieved by Random Forest models. There are six sets of models. `Real-Data': these are random forest models trained by  reducing the real data; `\tool': these are random forest models trained by reducing the real data, but
substituting the reduced data by synthetic data generated by \tool; `GMM', `BN', `WPGAN', and `CTGAN': these are similar to the `\tool' models but obtained by substituting the reduced data by the four baselines respectively. The x-axis represents how much real data is used from 100\% down to 0\%.

If we only use real data, the F1 score drops from 0.99 down to 0.97 as the amount of data decreases. Clearly, with no real data, we are unable to train a model. When we substitute real data with that generated by the baselines, the performance drops even quicker, because they do a poor job of capturing the temporal and attribute dependence. 
 Even in the absence of any real data, data generated by \tool results in an F1 score of  0.97, where the drop in performance is only 2\%. That is, the model built with only synthetic data retains 98\% of the performance of the all real data trained model. 

\textbf{Task2: $bytes$ value forecasting.} follows a similar setup of experiments as Task1. Figure~\ref{fig:task2} shows the mean square error achieved by a neural network regression model. 
The plot shows that \tool and Bayesian network (BN) outperform the other three baseline models. 
Building a Bayesian network with domain knowledge typically performs better than GANs \cite{xu2019modeling}.

In our experiments, BN is optimized specifically for the $bytes$ sequential value.  However, \tool has two advantages over the Bayesian network. First, users do not need the domain knowledge required for Bayesian network implementation. Secondly, there is no inherent bias attributable to an expert unlike traditional Bayesian networks. Similar to the first task, the penalty for using only \tool generated data (with no real data) is low, an increase of 13\% in the mean square error.

\textbf{Observation 3:} \emph{
Compared to BN, \tool performs better on task1 and as well on task2 without requiring any domain knowledge.
}

\textbf{Observation 4:} \emph{
Even with 0\% real data, \tool models task1 and task2 with only a small drop in accuracy.
}

%% file: conclusion.tex
\section{Conclusion and Future Work}
This paper presents the design and implementation of
\tool, a novel, flexible and robust approach to learn the distribution of complex multivariate time-series data distributions. Compared to existing approaches,
\tool is novel in several aspects. First, \tool learns the joint distribution over both temporal dependency and attribute dependency. Second, \tool is able to  generate data with any combination of continuous and discrete attributes. Furthermore, our architecture specifically supports generation of IP addresses and port numbers, 
which makes it particularly suitable for network traffic data.
We perform a thorough evaluation of \tool comparing it with four baselines using several performance measures as well as on 
two cybersecurity machine learning tasks. 

In the future, we plan to conduct a general and robust privacy preserving evaluation of the generated data. In particular, we plan to empirically validate privacy of training data, that is, no training data is leaked in the generated synthetic data by conducting privacy attacks, such as membership inference attacks \cite{shokri2017membership}, and also ensuring there is no training data memorization in our model \cite{carlini2019secret}. Other future work includes: 
(1) Experimenting with larger filters to validate modeling of longer term temporal dependency in training data.
(2) Generate anomalous (attack) data in addition to normal data.
(3) explore the best updating rate for re-learning the data synthesizer on historical data $\textbf{D}_{historical}$ on an ongoing basis; (4) conduct more semantic or statistic checking with regards to the fungibility of synthetic data with real data; and (5) support training with and generation of IPv6 addresses.

\textbf{Acknowledgment} This work is supported in part by the Commonwealth Cyber Initiative (CCI) and US NSF grant DGE-1545362. Any opinions, findings, and conclusions or recommendations expressed in this material are those of the author(s) and do not necessarily reflect the views of the sponsors.